\documentclass[preprint,12pt]{elsarticle}
\usepackage{lineno}
\usepackage{bm}
\usepackage{amsfonts}
\usepackage{amssymb}
\usepackage{amsmath}
\usepackage{isomath}
\usepackage{pgf}
\usepackage{adjustbox}
\usepackage{svg}
\usepackage{tabularx}

\usepackage{scalerel,stackengine}
\stackMath
\newcommand\reallywidehat[1]{%
\savestack{\tmpbox}{\stretchto{%
  \scaleto{%
    \scalerel*[\widthof{\ensuremath{#1}}]{\kern-.6pt\bigwedge\kern-.6pt}%
    {\rule[-\textheight/2]{1ex}{\textheight}}
  }{\textheight}%
}{0.5ex}}%
\stackon[1pt]{#1}{\tmpbox}%
}

\usepackage{stmaryrd} 
\usepackage{graphicx}
\graphicspath{{Figures/}}
\usepackage{float}
\usepackage{algorithm}
\usepackage{algpseudocode}
\usepackage{caption}
\usepackage{subcaption}
\usepackage[usenames]{xcolor}
\usepackage{multirow}
\usepackage{natbib}
\modulolinenumbers[5]
\usepackage[left = 2cm, right=2.5cm, top=2cm, bottom =2cm]{geometry}
\usepackage{hyperref}
\begin{document}
\begin{frontmatter}

\title{ Learning the Constitutive Behavior of Materials
via Neural Operators and Causal Attention: Case Studies in Plasticity and Damage}


\author{Rishabh Arora$^{1,*}$, Lisa Scheunemann$^{1}$, Tim Brepols$^{1}$, Shahed Rezaei$^{2,*}$}
\address{$^1$Institute of Applied Mechanics, \\ RWTH Aachen University, Mies-van-der-Rohe-Str. 1, D-52074 Aachen, Germany}
\address{$^2$ACCESS e.V., Intzestr. 5, D-52072 Aachen, Germany}
\address{$^*$ corresponding authors: rishabh.arora@ifam.rwth-aachen.de, s.rezaei@access-technology.de}

\begin{abstract}
Classical constitutive modeling of path-dependent inelastic materials
relies on internal state variables whose evolution equations must be
postulated based on domain knowledge and calibrated against experimental data.
However, in many practical settings, the relevant internal variables
are typically not measurable in experiments, and the constitutive response must be inferred entirely from measured strain-stress data without any
prior knowledge of the material's internal state. We propose a data-driven constitutive modeling framework based on the concept of a material operator, which treats a deforming material as a functional mapping from its entire strain history to the corresponding stress response. In contrast to traditional autoregressive or recurrent formulations, the model is trained directly on full loading paths as function-to-function mappings, predicting complete stress trajectories in a single parallel forward pass. Temporal path dependence is enforced through a causally masked attention mechanism embedded within the operator, which restricts the model's attention to past material states while preserving computational parallelizability. Spectral convolutions provide discretization-invariant representations in the frequency domain, while causal attention captures highly adaptive, non-local history dependence. Furthermore, sinusoidal activation functions are used to resolve the strong nonlinear transitions inherent in inelastic regimes. The framework is evaluated across multidimensional, rate-independent material models exhibiting complex phenomena, with an emphasis on nonlinear plasticity and ductile damage accumulation. The results demonstrate accurate and robust predictions of irreversible deformation mechanisms while simultaneously achieving resolution invariance and excellent parallel efficiency.

\end{abstract} 

\end{frontmatter}

\section*{Nomenclature}
\begin{tabularx}{\textwidth}{@{}l X@{}}
\textbf{Abbreviation} & \textbf{Meaning} \\[2pt]
\hline \\[-8pt]
ANN   & Artificial Neural Network \\
CANN  & Constitutive Artificial Neural Network \\
CP    & Crystal Plasticity \\
FCNN  & Fully Connected Neural Network \\
FEM   & Finite Element Method \\
FFNN  & Feed Forward Neural Network \\
FFT   & Fast Fourier Transform \\
FNO   & Fourier Neural Operator \\
GP    & Gaussian-Process \\
GRU   & Gated Recurrent Unit \\
ISV   & Internal State Variable \\
KKT   & Karush--Kuhn--Tucker \\
LN    & Layer Normalization \\
LSTM  & Long Short-Term Memory \\
MLP   & Multilayer Perceptron \\
MSE   & Mean Squared Error \\
PDE   & Partial Differential Equation \\
PINN  & Physics-Informed Neural Network \\
RNN   & Recurrent Neural Network \\
RVE   & Representative Volume Element \\
TANN  & Thermodynamics-based Artificial Neural Network \\
SIREN & Sinusoidal Representation Network \\
TPE   & Tree-structured Parzen Estimator \
\end{tabularx}

\section{Introduction} 
Constitutive models lie at the heart of every computational solid mechanics simulation. They define the relationship between deformation and stress that is evaluated at every integration point during every time step, and the fidelity of a finite element analysis is therefore strongly dependent on the fidelity of the constitutive law it employs. For path-dependent inelastic materials that exhibit elastoplasticity, damage, and coupled plastic–damage behavior, formulating such a law remains a central challenge in computational solid mechanics. The defining feature of these materials is memory, which means that the stress at any instant depends not only on the current strain but also on the entire prior deformation history. This history dependence cannot be captured by a single algebraic stress–strain relationship and is usually encoded through internal state variables whose evolution traces the irreversible processes within the material \cite{Coleman1967,Simo2006,lubliner2008plasticity}. Classical phenomenological models describe such behavior through these internal state variables, governed by evolution equations together with formulation of suitable yield and damage surfaces \cite{DeSouzaNeto2008}. This framework is powerful and physically transparent, but constructing a specific model is demanding. The task becomes considerably harder when several internal mechanisms are active simultaneously. Plastic hardening, damage growth, and their mutual coupling interact in ways that are difficult even for an experienced modeler to anticipate. Damage can, for example, degrade the effective stiffness that drives plastic flow, while plastic deformation in turn drives the energy release that fuels damage. The two mechanisms may evolve together under loading paths involving arbitrary sequences of loading, unloading, and reversal \cite{lemaitre1985coupled}.

These difficulties have motivated data-driven constitutive modeling, which aims to learn the stress–strain response directly from experimental or simulation data without prescribing the functional form of the constitutive law in advance. The use of neural networks to represent material behavior dates back to the pioneering work of \citet{Ghaboussi1991,Ghaboussi1999}, who first proposed learning constitutive relations directly from data with multilayer perceptron (MLP), exploiting the universal approximation property of feedforward networks \cite{Hornik1989}. This idea was quickly taken up across a range of material classes and, in the years since, has grown into a substantial and internationally distributed body of work. \citet{Furukawa1998} developed an implicit neural viscoplastic model in which the network predicts the rates of the viscoplastic strain and the internal variables, extending the approach to rate-dependent behavior. \citet{fernandez2020} applied artificial neural networks as a direct data-driven surrogate for grain-boundary interface mechanics, learning the constitutive behavior of grain boundaries from data rather than deriving it from a prescribed physical model.  \citet{Zhang2020} demonstrated the capability of neural networks to effectively capture path-dependent material behavior by modeling von Mises plasticity with isotropic hardening. To capture loading history without an explicit window, \citet{HUANG2020113008} developed a machine learning-driven material modeling framework in which plasticity is represented through a Proper Orthogonal Decomposition, with the accumulated absolute strain serving as the history variable driving the reduced-order plastic response. \citet{fernandez2021anisotropic} combined material theory with data-driven calibration to construct anisotropic hyperelastic constitutive models for finite deformations, applying the approach to cubic lattice metamaterials. \citet{Mianroodi2021} trained a convolutional network to map an image of a nanoporous microstructure directly to its elasticity tensor computed via molecular statics.


Utilizing convolutional neural networks, \citet{Frankel_2020} successfully predicted the spatiotemporal evolution of the stress field based on the initial microstructural features and applied external loading conditions. To accelerate computationally expensive crystal plasticity finite element simulations, \citet{ALI2019205} developed Feed Forward Neural Network (FFNN)-based surrogate models focused on capturing the homogenized response of the Representative Volume Element (RVE). Further work in this field is reviewed in detail in \cite{Fuhg2024,hussain2026machine,Dornheim2023,rosenkranz2023comparative}.

 These contributions established the central premise of the field, that a sufficiently expressive network can encode a stress–strain relationship without a prescribed functional form, and formed the basis for much subsequent work. Despite their influence, these models shared several limitations that have shaped the research directions since. Their reliance on a fixed window of preceding time steps ties the learned response to the specific temporal sampling of the training data, which leads to accuracy degradation when the model is used at a different time step size. 
 


A fundamentally different strategy is the model-free data-driven computing paradigm, which dispenses with a constitutive model altogether rather than learning one via a neural network. \citet{Kirchdoerfer2016} and \citet{EGGERSMANN201981} introduced this approach for linear elastic and inelastic material, respectively. The study examined materials with memory (using the full deformation history), differential materials (using only a short recent history), and history variables (using ad-hoc auxiliary variables). Each representation involves a natural trade-off that continues to motivate further development. The full-history representation offers the most complete description of memory but increases the dataset's dimensionality, favoring short-to-moderate horizon problems, while the history-variable representation is more scalable but leaves the question open of how such variables might be systematically selected or optimized.

A substantial body of work has since sought to restore physical consistency for inelastic materials by embedding known structure directly into the learned model. One direction enforces the laws of thermodynamics by construction (TANN by \citet{MASI2021104277,Masi2020}), so that the predicted response is admissible by design rather than only during training. Another line of work by \citet{VLASSIS2021113695} instead builds thermodynamic consistency into a thermodynamics-informed neural network. This approach uses deep network predictions to evolve interpretable components like the yield surface and plastic flow rule, which depend on internal variables. To solve material-specific differential equations, \citet{HAGHIGHAT2023105828} detailed an alternative methodology that uses Physics-Informed Neural Network (PINN) \cite{RAISSI2019} as its foundational framework. By training a neural network to predict the Cholesky factor of the tangent stiffness matrix, \citet{XU2021110072} established an alternative method for stress prediction. Ultimately, this construction improves numerical stability in finite element simulations by weakly enforcing strain energy convexity. A data-driven, multiscale approach utilizing physics-constrained neural networks was introduced by \citet{Kalina2023} to model finite strain hyperelasticity. Related research by \citet{KLEIN2022104703} also explores the development of constitutive models using polyconvex neural networks. \citet{Rezaei2024} introduced COMM-PINN, a physics-informed neural network that instantly resolves nonlinear, path-dependent constitutive equations of materials and satisfies thermodynamic constraints during the training time without relying on initial training data. The work by \citet{Rosenkranz2024} presents a physics-augmented neural network model for nonlinear viscoelastic materials that is thermodynamically consistent by construction (built on free energy and dissipation potentials via generalized standard materials). \citet{Aldakheel2025} embedded the governing equations of brittle and ductile fracture in elastic-viscoplastic materials directly into a feedforward network's architecture, enforcing thermodynamic consistency by construction. \citet{Asad2023} proposed a mechanics-informed training procedure that constrains the learned model to satisfy dynamic and material stability, while \citet{Roy2023} embedded the governing equations of von Mises plasticity directly into a physics-informed network for plane-strain boundary value problems. 

A second direction seeks not a black-box surrogate but an interpretable method. For example, sparse-regression approaches over a library of candidate terms discover the governing equations directly \cite{Brunton2016}, an idea recently extended by pairing multiple sparse-regression algorithms with model-selection criteria to automate constitutive model discovery \cite{URREAQUINTERO2026118551}. The Constitutive Artificial Neural Network (CANN) framework of \citet{Linka2021} and \citet{LinkaKuhl2023} structures the network so that its architecture embeds kinematic and thermodynamic constraints and its weights carry direct physical meaning, turning model selection into automated model discovery. These ideas have since been extended to inelastic behavior \citep{Holthusen2024} and to anisotropic finite-strain viscoelasticity \citep{Abdolazizi2024}. While these physics-structured models are accurate, interpretable, and data-efficient, they come with a trade-off. They require reintroducing the exact domain assumptions that purely data-driven methods set out to avoid, such as a specific free-energy form, a chosen set of invariants, or an assumed family of internal variables. When internal state variables are discovered automatically, through the encoder-decoder construction stated in the work of \citet{masi2023evolution}, the user must still prescribe an underlying thermodynamic potential and dissipation format. These approaches are therefore most effective when the underlying physical structure is at least partially known. Modeling path-dependent behavior accurately when no such structure is prescribed, when the constitutive response, including its internal memory, must be inferred solely from measured strain-stress histories, remains a challenging and largely open problem, and is the setting addressed in this work.

Parallel to the development of physics-embedded approaches, a longstanding purely data-driven strategy for path-dependent materials has relied on sequence-modeling architectures to capture history dependence without explicit internal state variables. FFNN with explicit history windows \cite{Ghaboussi1991,Noor2026} maps a finite window of past strain–stress pairs to the next stress increment in an autoregressive fashion: at every time step, the network first predicts the new stress ${\boldsymbol{\sigma}}_{n+1}$ based on the most recent $k$ strain–stress observations and the new strain increment, the prediction is then fed back into the history buffer, and the procedure is repeated. Recurrent neural network (RNN) and its gated variants avoid an explicit input window by encoding loading memory in an evolving hidden state ${h}_t$, which is updated sequentially one step at a time as the strain trajectory is consumed; the gating mechanism of \citet{Cho2014} in particular lets the network learn what to retain and what to discard from this hidden state at every step, rather than carrying the full history forward unfiltered. \citet{Gorji2020} established this paradigm for path-dependent plasticity by training gated recurrent unit networks to reproduce the response of an anisotropic plasticity model with homogeneous anisotropic hardening under complex, non-monotonic plane-stress loading. They also showed that the resulting surrogate generalizes to loading paths not seen during training. \citet{Chen2021} applied the same recurrent structure to viscoelastic materials, training on synthetic strain-stress sequences generated from a classical viscoelastic constitutive law and showing that the RNN captures the rate-dependent response without a prescribed Prony-series or relaxation-spectrum form, generalizing from simple linear and step strain inputs to unseen quadratic loading paths. \citet{Mozaffar2019} showed that recurrent neural networks, trained on representative volume element simulations, can predict path-dependent plastic response directly from strain histories without postulating an explicit yield surface. \citet{Bonatti2022CP} scaled the recurrent approach to full crystal plasticity, training an RNN surrogate directly on fast-Fourier-transform-based crystal plasticity simulations to replace the costly micromechanical solve at each macroscopic integration point. \citet{Maia2023} took a hybrid route, embedding a classical constitutive model inside the recurrent cell itself so that the network's hidden state coincides with genuine physical internal variables, combining the flexibility of a data-driven surrogate with the interpretability of a mechanistic one. Both paradigms thus process the loading path temporally: the strain trajectory is presented as a stream of values, and path dependence is realized either through an explicit recursion over a buffer of recent observations (FFNN) or through an implicit recursion over a hidden state (RNN). Operator-based extensions such as the history-aware neural operator in recent studies by \citet{Guo2025} retain the autoregressive structure of FFNN-style surrogates while replacing the underlying network with a Fourier Neural Operator (FNO), inheriting some discretization invariance from the spectral backbone but still operating in the next-step-prediction mode. A consequence of this temporal processing is that the model representation is tied to the temporal discretization of the training data: a network trained on loading paths sampled at one temporal resolution typically degrades when evaluated at coarser or finer resolutions, violating the principle that a rate-independent constitutive response should depend on the loading path itself rather than on how it is sampled. This loss of self-consistency has been documented for standard RNN architectures in the work by \citet{Bonatti2022}. Achieving discretization invariance without sacrificing predictive accuracy on complex loading paths is the central methodological objective of the present work. 

\begin{figure}[H] 
  \centering
  \includegraphics[width=0.99\linewidth]{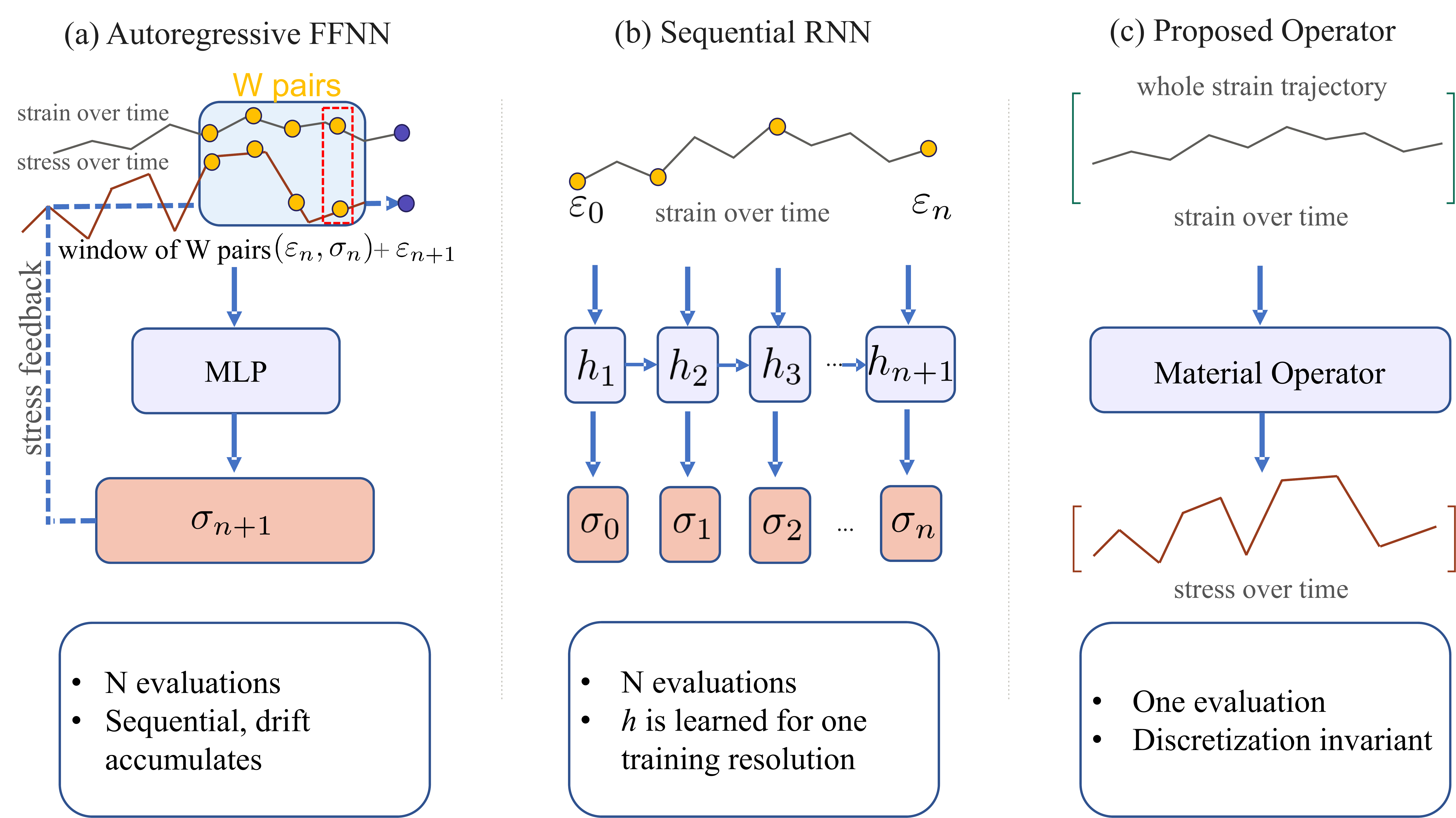}
  \caption{ Three paradigms for data-driven path-dependent constitutive modeling at inference time (a) Autoregressive FFNN (b) Sequential RNN-GRU (c) Proposed material operator }
  \label{fig:paradigms}
\end{figure}

We propose a constitutive modeling framework that departs from both the autoregressive and the sequential-recurrent paradigms by treating the entire loading path as a single function input. Given a discretely sampled strain trajectory $\boldsymbol{\varepsilon} = (\boldsymbol{\varepsilon}_0, \boldsymbol{\varepsilon}_1, \ldots, \boldsymbol{\varepsilon}_{n})$, the proposed operator produces the full stress trajectory ${\boldsymbol{\sigma}} = ({\boldsymbol{\sigma}}_0, {\boldsymbol{\sigma}}_1, \ldots, {\boldsymbol{\sigma}}_{n})$ in a single parallel forward pass. The temporal index acts as a spatial-like coordinate over which the operator performs spectral convolution and self-attention.  Within each block, a strict causal mask restricts every attention layer so that its output at a given temporal position depends on past and present strain values. The spectral convolution that follows operates globally over the temporal axis to provide resolution-invariant mixing of these causally-attended features. Figure~\ref{fig:paradigms} presents the three paradigms: autoregressive FFNN, sequential RNN-GRU (Gated Recurrent Unit), and the proposed sequence-to-sequence operator, highlighting how the input is presented and how path dependence is realized in each case during inference time.

Neural operators generalize neural networks to map between function spaces \cite{Li2021,Kovachki2023}. For time-dependent PDEs, they have most often been applied autoregressively in time. The operator advances the full spatial field by one increment and is rolled forward step by step over the temporal horizon \cite{yamamazaki2025finite,najianasl2026physics,Li2021,wen2022ufno}. Each such evaluation is a complete function-to-function map over space but not in time. We depart from this step-wise view. Rather than treating time as a stream advanced increment by increment, we treat the entire loading path as a single function-valued input and learn a direct mapping to the corresponding stress trajectory. The FNO implements this through spectral convolution in the frequency domain, retaining a finite set of modes \cite{Li2021}. This formulation enables discretization-invariant behavior, since the learned filters are tied to physical wavenumbers instead of discrete time indices. As a result, the same model can be evaluated on loading paths sampled at different temporal resolutions without retraining. The main focus of our work is how it stands apart from earlier neural operators. In the recent history-aware neural operator for constitutive modeling \cite{Guo2025}, the system still moves forward step by step by predicting one increment from the previous one. This step-wise recursion makes those models sensitive to the temporal step size. The proposed material operator is instead applied once to the entire loading path, mapping the complete strain trajectory to the full stress trajectory in a single parallel pass without any rollout. This setup lets us compare, on a common footing, the three ways of representing path dependence that structure this paper as shown in Figure \ref{fig:paradigms}.

To resolve the sharp features of inelastic response, the architecture combines the spectral structure of the FNO \cite{Li2021} with the sinusoidal representation networks
(SIREN) of \citet{Sitzmann2020}, whose periodic activations of the form
$\sin(\omega_0 {W}{x} + {b})$ are particularly well suited
to representing the sharp transitions characteristic of yield onset,
elastic-to-plastic switching, and damage initiation. Standard non-periodic
activations instead exhibit a spectral bias toward smooth, low-frequency
signals \cite{Rahaman2019}. The choice of
$\omega_0$ controls the spanned frequency spectrum, enabling sharp transitions
to be captured without an excessive number of layers or parameters
\cite{Sitzmann2020}. Finally, the self-attention mechanism of \citet{Vaswani2017} is adopted, with strict causal masking. This provides adaptive, content-dependent weighting of past values (i.e.\ strain) without compromising the directional flow of mechanical history.

We develop an FNO-SIREN architecture that integrates spectral operator learning with sinusoidal representation networks for data-driven constitutive modeling of rate-independent path-dependent materials, and we further enhance it with a self-attention mechanism for history weighting. We demonstrate that the proposed architecture achieves strong path-family generalization, in that a model trained exclusively on Gaussian-Process loading paths accurately predicts the stress response on unseen style distribution zig-zag and sinusoidal paths. We validate the framework on three benchmark problems of increasing complexity. These consist of a one-dimensional elastoplasticity material model with nonlinear isotropic hardening, a one-dimensional coupled plastic-damage material model, and a two-dimensional elastoplasticity material model.

\section{Thermodynamics-based material modeling}
For rate-independent inelastic solids under isothermal conditions, the constitutive response is conventionally formulated within the framework of irreversible thermodynamics with internal state variables \cite{Coleman1967,Simo2006}. Let $\boldsymbol{\varepsilon}(t)$ denote the strain at time $t$ and $\mathbf{z}(t) = (\mathbf{z_1}, \ldots, \mathbf{z_n})$ a collection of Internal State Variables (ISVs) encoding the irreversible material memory representing plastic strain, hardening parameters, or damage variables. The Helmholtz free energy is postulated as $\Psi = \Psi(\boldsymbol{\varepsilon}, \mathbf{z})$, from which the Cauchy stress and thermodynamic conjugate forces follow as
\begin{equation}
\boldsymbol{\sigma} = \frac{\partial \Psi}{\partial \boldsymbol{\varepsilon}}, \qquad \mathbf{Y} = -\frac{\partial \Psi}{\partial \mathbf{z}}.
\end{equation}
The remaining part of the Clausius–Duhem inequality $\mathcal{D} = \mathbf{Y} \cdot \dot{\mathbf{z}} \geq 0$ constrains the evolution of the ISVs, typically prescribed through a rate equation $\dot{\mathbf{z}} = \mathbf{f}(\mathbf{Y}, \boldsymbol{\sigma}, \mathbf{z})$ together with admissibility conditions, for example, for rate-independent plasticity and damage. The resulting constitutive law $\boldsymbol{\sigma}(t) = \boldsymbol{\sigma}(\boldsymbol{\varepsilon}(t), \mathbf{z}(t))$ is implicitly path-dependent. The stress at any instant depends on the entire prior strain history through the evolution of $\mathbf{z}$.

In the remainder of this paper we evaluate the proposed data-driven framework on three benchmark constitutive models of increasing complexity. The first is a one-dimensional elastoplastic model with nonlinear isotropic hardening, which serves as the baseline scalar setting in which all three model architectures are compared. The second is a one-dimensional coupled damage-plasticity model, which couples plastic flow with progressive isotropic damage and tests whether the architectural advantages observed in pure plasticity carry over to a regime with two simultaneous irreversible processes. The third is a two-dimensional plane-strain elastoplastic model, which generalizes the first benchmark to a multi-component setting where the surrogate must simultaneously predict three stress components from three strain components and learn the coupling between normal and shear directions.

\subsection{One-dimensional elastoplastic material model } 
\label{subsec: 1-d elasto}
We consider a one-dimensional elastoplastic model with nonlinear isotropic hardening of saturation type. The total strain is additively decomposed as $\varepsilon = \varepsilon^e + \varepsilon^p$, with the stress given by $\sigma = E \varepsilon^e$ and the yield function
\begin{equation*}
    \phi(\sigma, \xi^p) = |\sigma| - \big(\sigma_y + h_1 (1 - e^{-h_2 \xi^p}) \big),
\end{equation*}
where $\xi^p$ is the accumulated plastic strain and the hardening term saturates exponentially with rate $h_2$ and asymptotic amplitude $h_1$.The associated flow rule is given by $\dot{\varepsilon}^p = \dot{\gamma}\,\mathrm{sign}(\sigma)$ and $\dot{\xi}^p = \dot{\gamma}$, where the plastic multiplier $\dot{\gamma}$ satisfies the Karush--Kuhn--Tucker (KKT) conditions $\dot{\gamma} \geq 0$, $ \phi \leq 0$, $\dot{\gamma}\,\phi = 0.$ The constants used throughout this section are $E = 3.0$ MPa, $\sigma_y = 0.6$ MPa, $h_1 = 0.4$ MPa, and $h_2 = 10.0$. Reference stress histories are generated by a return-mapping algorithm \cite{Simo2006}. The algorithm is stated in the work of \citet{Rezaei2024}.

\subsection{One-dimensional coupled damage--plasticity material model} \label{subsec:1d_damage}
To probe a setting in which two irreversible mechanisms evolve simultaneously, we adopt a one-dimensional coupled isotropic two-surface damage--plasticity model defined in the work of \citet{Brepols2017,Brepols2018}. We consider here a purely local formulation of damage, i.e., without a gradient-extension. The model introduces a scalar isotropic damage variable $D \in [0, 1)$ that enters the effective material stiffness through the degradation function

$$f(D) = 1 - D,$$

such that the stiffness is ultimately degraded by $f(D)^2$, and the Cauchy stress is expressed in terms of the elastic strain $\varepsilon^e = \varepsilon - \varepsilon^p$ as

$$\sigma = f(D)^2 E \varepsilon^e.$$

The squared damage factor reflects the dual reduction of effective load-bearing area in both the stress--strain relation and the conjugate thermodynamic forces, consistent with the energy-equivalence interpretation in continuum damage mechanics. In addition to $\varepsilon^p$ and $D$, which are themselves internal variables governing plastic and damage evolution, two further internal state variables track the hardening response of the two irreversible mechanisms: $\xi^p$ for isotropic plastic hardening, evolving in the effective configuration, and $\xi^d$ for damage hardening.

The plastic yield function is
$$
F_p(\sigma, \xi^p, D) = |\sigma| - f(D)\sigma_0 - f(D)^2 h_p \xi^p,
$$
where $\sigma_0$ is the initial yield stress and $h_p$ represents the linear plastic hardening modulus. The associated flow rule is $\dot{\varepsilon}^p = \dot{\gamma}_p \, \mathrm{sign}(\sigma)$, and the plastic hardening variable evolves as \mbox{\(\dot{\xi}^p = \dot{\gamma}_p / f(D)\)}, so that hardening progresses faster as the material softens through damage.

Damage growth is governed by the elastic energy release rate
$$
Y(\varepsilon^e, \xi^p, D) = f(D) \big( E(\varepsilon^e)^2 + h_p(\xi^p)^2 \big),
$$
together with the damage activation criterion
$$
F_d(Y, \xi^d) = Y - \big( Y_0 + r_d \xi^d \big),
$$
where $Y_0$ is the threshold for damage initiation and $r_d$ is the linear damage hardening modulus. The damage variable evolves associatively as $\dot{D} = \dot{\gamma}_d$, while the damage hardening variable follows the non-associative rule $\dot{\xi}^d = \dot{\gamma}_d \big(1 + \tfrac{s_d}{r_d} q_d\big)$ with $q_d = r_d \xi^d$. Upon backward-Euler time discretization, these evolution equations are integrated over each time increment $\Delta t$ in terms of the incremental multiplier $\Delta \gamma_d = \dot{\gamma} \Delta t \geq 0$, yielding the closed-form algorithmic update
$$
\Delta D = \Delta\gamma_d, \qquad \Delta\xi^d = \frac{\Delta\gamma_d}{1 + s_d\,\Delta\gamma_d},
$$
which is the form used in the local Newton solve described below, with $s_d$ saturating the damage hardening rate at large increments. Both plastic and damage multipliers satisfy independent Karush--Kuhn--Tucker conditions, stated here in discrete form as $F_p \leq 0$, $\Delta\gamma_p \geq 0$, $\Delta\gamma_p F_p = 0$ and $F_d \leq 0$, $\Delta\gamma_d \geq 0$, $\Delta\gamma_d F_d = 0$.

The constitutive response is computed by a stress update that branches into four cases depending on the activity of the plastic and damage surfaces: a purely elastic case, an elastoplastic-only case, an elastic-damage-only case, and a coupled case in which both $F_p = 0$ and $F_d = 0$. The coupled case requires the simultaneous solution of two consistency equations in the two multipliers $\Delta\gamma_p$ and $\Delta\gamma_d$, which is performed by a Newton iteration on the $2 \times 2$ residual system; further details can be found in \cite{Brepols2017,Brepols2018}. Parameter values used in the present study are $E = 3.0$ MPa, $\sigma_0 = 0.6$ MPa, $h_p = 0.4$ MPa, $Y_0 = 0.15$ MPa, $r_d = 0.5$ MPa, $s_d = 0.05$.

\subsection{Two-dimensional plane-strain elastoplastic material model} \label{subsec:2d_elasto}
We finally extend the evaluation to a two-dimensional plane-strain setting, where the surrogate must simultaneously predict three stress components from three strain components throughout the loading history. For the multi-axial extension, we adopt classical $J_2$ (von Mises) plasticity, in which yielding is governed by the second invariant of the deviatoric stress tensor $\mathbf{s}$,

$$J_2 = \frac{1}{2}\,\mathbf{s}:\mathbf{s},$$

rather than by the normal and shear stress components individually. The associated von Mises equivalent stress is $q = \sqrt{3 J_2}$. The added difficulty is structural rather than just dimensional: the trained model must learn the coupling between normal and shear components through the deviatoric structure of the $J_2$ flow rule.

We consider rate-independent $J_2$ plasticity with nonlinear isotropic hardening of the same saturation form used in Section \ref{subsec: 1-d elasto}, generalized to two-dimensional plane strain. The total strain tensor is decomposed additively into elastic and plastic parts, $\boldsymbol{\varepsilon} = \boldsymbol{\varepsilon}^e + \boldsymbol{\varepsilon}^p$, and the Cauchy stress follows Hooke's law:
$$
\boldsymbol{\sigma} = \lambda \mathrm{tr}(\boldsymbol{\varepsilon}^e) \mathbf{I} + 2\mu \boldsymbol{\varepsilon}^e,
$$
with Lamé parameters $\lambda = E\nu/[(1+\nu)(1-2\nu)]$ and $\mu = E/[2(1+\nu)]$. The out-of-plane normal strain is constrained by plane-strain kinematics, $\varepsilon_{zz}^e = -(\varepsilon_{xx}^p + \varepsilon_{yy}^p)$, so the out-of-plane normal stress $\sigma_{zz}$ is non-zero and contributes to the deviatoric response. The deviatoric stress is $\mathbf{s} = \boldsymbol{\sigma} - \tfrac{1}{3}\mathrm{tr}(\boldsymbol{\sigma})\,\mathbf{I}$, and the equivalent von Mises stress is
$$
q(\mathbf{s}) = \sqrt{\tfrac{3}{2} \mathbf{s}:\mathbf{s}}.
$$
Plastic yielding is governed by the $J_2$ yield function
$$
\phi(q, \xi^p) = q - \big(\sigma_y + h_1 (1 - e^{-h_2 \xi^p})\big),
$$
with $\xi^p$ being the accumulated plastic strain. The associated flow rule defines the plastic strain rate as $\dot{\boldsymbol{\varepsilon}}^p = \dot{\gamma}\,\mathbf{n}$ and the equivalent plastic strain rate as $\dot{\xi}^p = \dot{\gamma}$. In this formulation, the flow direction is given by $\mathbf{n} = \frac{3}{2}\,\mathbf{s}/q$, and the non-negative plastic multiplier ($\dot{\gamma} \geq 0$) enforces the Karush--Kuhn--Tucker conditions, namely $\phi \leq 0$ and $\dot{\gamma}\phi = 0$. The material constants are inherited from Section~\ref{subsec: 1-d elasto} ($E = 3.0$ MPa, $\sigma_y = 0.6$ MPa, $h_1 = 0.4$ MPa, $h_2 = 10.0$) and the plane-strain setting is closed by $\nu = 0.3$. Reference responses are computed by a return-mapping algorithm with a Newton iteration on the consistency residual, following the standard procedure (see also \cite{Simo2006}).

The surrogate operates on three in-plane components: the strain history is represented as $\boldsymbol{\varepsilon}(t) = (\varepsilon_{xx}, \varepsilon_{yy}, \varepsilon_{xy})$ and the stress history as $\boldsymbol{\sigma}(t) = (\sigma_{xx}, \sigma_{yy}, \sigma_{xy})$, so the input and output of the operator are both functions taking values in $\mathbb{R}^3$ over the temporal axis.

\section{Data-driven constitutive modeling of path-dependent materials}
This section develops the proposed framework for data-driven constitutive modeling of path-dependent materials that are rate-independent. We review three paradigms, which are FFNN with explicit history windows, RNN-GRU, and, as the proposed alternative, a sequence-to-sequence operator built on a FNO backbone with sinusoidal representation layers and causal self-attention. 

The objective of data-driven constitutive modeling is to learn implicit history dependence directly from observed strain–stress data, without committing to a specific functional form. In continuous-time form, the target mapping is a \textit{causal operator}
\begin{equation}
\mathcal{G} : \boldsymbol{\varepsilon}(\cdot)\big|_{[0,t]} \;\longmapsto\; \boldsymbol{\sigma}(t),
\end{equation}
which maps the entire strain history up to time $t$ to the current stress, with the causality requirement that $\boldsymbol{\sigma}(t)$ should not depend on $\boldsymbol{\varepsilon}(\tau)$ for $\tau > t$. In practice, measurements are available at discrete instants $t_0 < t_1 < \ldots < t_{n}$, and the discrete counterpart maps the full sampled strain trajectory $(\boldsymbol{\varepsilon}_0, \boldsymbol{\varepsilon}_1, \ldots, \boldsymbol{\varepsilon}_{n})$ to the full sampled stress trajectory $(\boldsymbol{\sigma}_0, \boldsymbol{\sigma}_1, \ldots, \boldsymbol{\sigma}_{n})$ .

The three neural-network paradigms reviewed and developed in the remainder of this section differ in how they approximate this causal operator. Each makes a different structural choice about (i) what is presented to the network as input, (ii) how path dependence is enforced, and (iii) how predictions are produced at inference. Figure 1 contrasts these choices schematically.

\subsection{Sequence-to-sequence operator: a different way to present the loading path }

The conventional approach to path-dependent surrogate modeling is to process the loading path (strain trajectory) as a stream of values, consumed one step at a time, and stress predictions are produced sequentially. The proposed framework departs from this convention. Instead of streaming the strain trajectory, we treat the entire discretely sampled loading path as a single function-valued input and learn an operator that produces the entire stress trajectory in one parallel forward pass:
\begin{equation}
{\boldsymbol{\sigma}} = \mathcal{G}(\boldsymbol{\varepsilon}), \qquad \boldsymbol{\varepsilon} = (\boldsymbol{\varepsilon}_0, \ldots, \boldsymbol{\varepsilon}_{n}), \quad {\boldsymbol{\sigma}} = ({\boldsymbol{\sigma}}_0, \ldots, {\boldsymbol{\sigma}}_{n})
\end{equation}
The temporal index plays the role of a spatial-like coordinate over which the operator acts; the loading path is, in effect, lifted into a one-dimensional function-space representation. Path dependence is the requirement that ${\boldsymbol{\sigma}}_i$ depend only on $\boldsymbol{\varepsilon}_0, \ldots, \boldsymbol{\varepsilon}_i$. It is no longer realized through iteration over a buffer or through a hidden-state recurrence. The proposed architecture is discussed further.

\subsubsection{Fourier Neural Operator backbone}
In the proposed architecture, each hidden operator layer $\{\mathcal{L}_\ell\}_{\ell=1}^{L}$ takes the form (\citet{Li2021}):
\begin{equation}
\mathcal{L}_\ell\mathbf{v}_{l-1}(x) = \rho\big( W_\ell
\mathbf{v}_{\ell-1}(x) + (\mathcal{K}_\ell \mathbf{v}_{\ell-1})(x) \big),
\end{equation}
where L is the total number of Fourier layers between the projection and lifting layers, $W_\ell$ is a pointwise linear map, $\mathcal{K}_\ell$ is the integral
kernel operator, and $\rho$ is a nonlinear activation function applied pointwise (meaning it acts independently on each component of the vector). We implement
$\mathcal{K}_\ell$ as a global convolution in the frequency domain (\citet{Li2021}):
\begin{equation}
(\mathcal{K}_\ell \mathbf{v}_{\ell-1})(x) = \mathcal{F}^{-1}\!\big[ R_\ell
\cdot \mathcal{F}[\mathbf{v}_{\ell-1}] \big](x),
\end{equation}
where $\mathcal{F}$ and $\mathcal{F}^{-1}$ denote the (fast) Fourier transform
and its inverse, and $R_\ell$ is a learnable complex-valued spectral filter
retaining only the lowest $M$ Fourier modes. Because $R_\ell$ indexes physical
wavenumbers rather than discrete grid points, the same trained filter applies
consistently across coarse and fine discretizations of the temporal axis,
provided the spectral content of the signal is adequately resolved
\cite{Kovachki2023}. This is the discretization invariance we require of our constitutive framework.

For the present application, the input function is the strain trajectory $\boldsymbol{\varepsilon}$, the domain $\mathcal{D}$ is the temporal index axis, and the output is the stress trajectory $\boldsymbol{\sigma}$. The FNO layer captures global temporal correlations within the loading path through its spectral convolution; the next two subsections explain our operator capacity to represent sharp transitions and to enforce causality.

\subsubsection{Sinusoidal representation layers}

To represent the sharp transitions characteristic of yield onset,
elastic-to-plastic switching, and damage initiation, we employ sinusoidal
representation layers (SIREN) \cite{Sitzmann2020}, in which every fully
connected layer uses a sine activation:
\begin{equation}
\textbf{h}_{\ell} = \sin\!\big( \omega_0 \, {U}_\ell \textbf{h}_{\ell-1} +
{b}_\ell \big),
\end{equation}
with ${U}_\ell$ as the weight matrix, and $\omega_0$ a frequency hyperparameter and the weights initialized to
preserve the variance of activations and gradients across layers.
 
We employ SIREN layers in three roles in the proposed architecture: (i) as the
lifting operator $\mathcal{P}$ that maps the scalar strain input
$\varepsilon(t)$ into the wider latent feature space, (ii) as the activation
function within the Fourier layers (FNO-SIREN) applied after combining the
spectral and local linear transformations, and (iii) as the projection
operator $\mathcal{Q}$ that maps the latent features back to the stress output
$\sigma(t)$. In these roles, SIREN acts pointwise along the temporal axis.

\subsubsection{Causal self-attention and the FNO-SIREN block}
\begin{figure}[H] 
  \centering
  \includegraphics[width=0.99\linewidth]{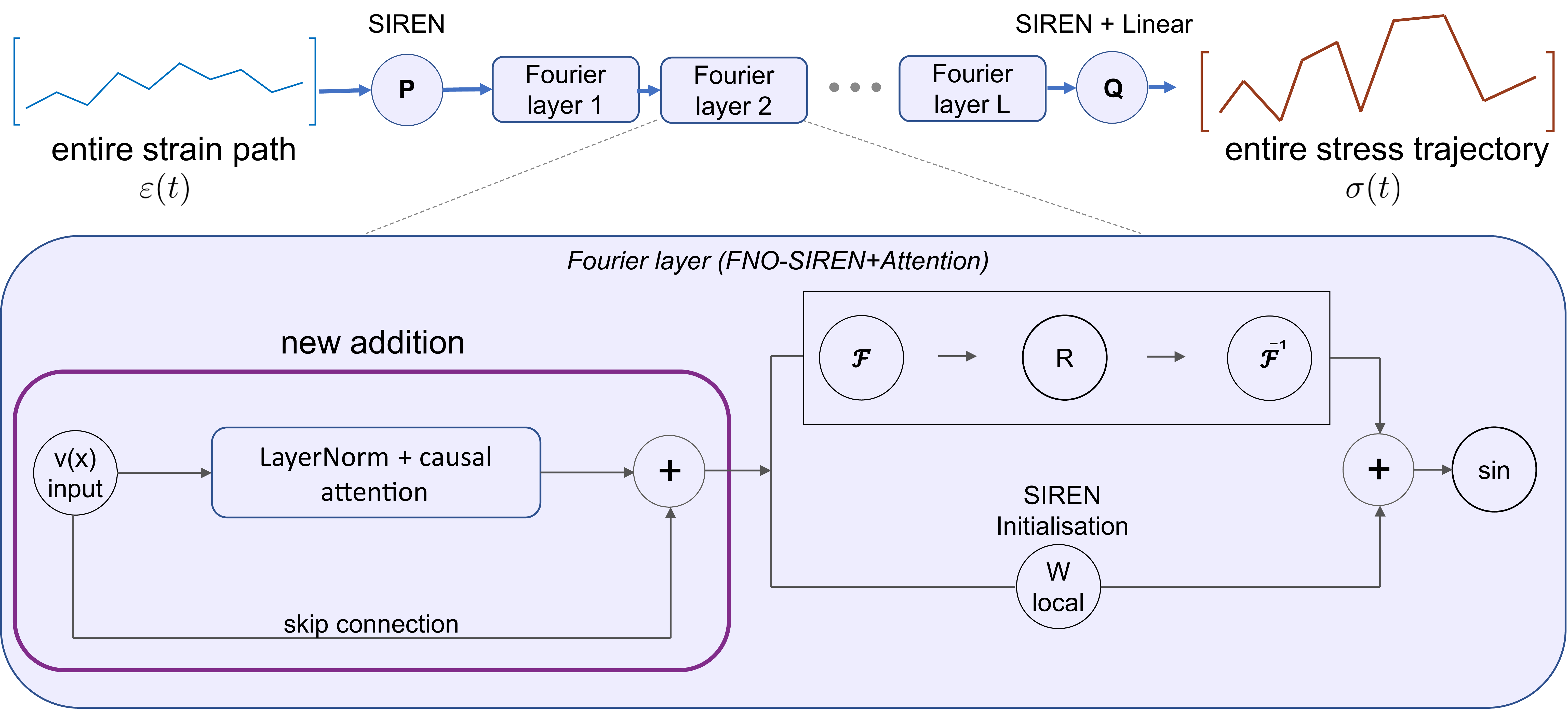}
  \caption{FNO-SIREN+Attention }
  \label{fig:FNO_SIREN_ATTENTION}
\end{figure}

\begin{figure}[H] 
  \centering
  \includegraphics[width=0.90\linewidth]{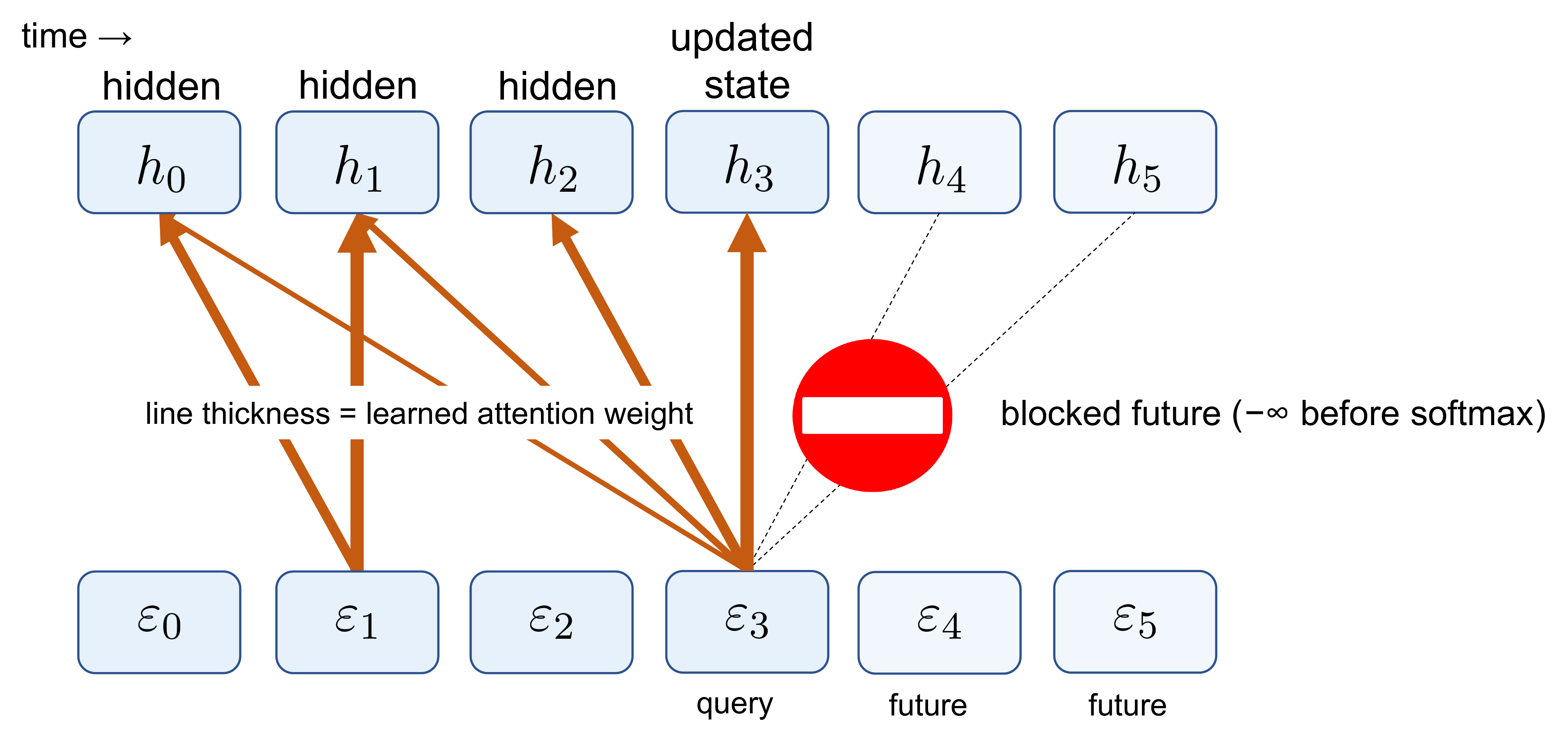}
  \caption{ Causal Attention Mechanism  }
  \label{fig:atten}
\end{figure}

While the spectral convolution captures global temporal correlations and the SIREN layers sharpen local features, both treat every position in the loading trajectory equally. In particular, the spectral convolution is non-causal. The Fourier transform couples each output position to both past and future strain values, so the predicted stress at a given instant is allowed to depend on strain values that occur later in the loading path. This violates a basic physical requirement of a path-dependent constitutive law, namely that the stress at any instant may depend only on the prior strain history and not on future deformation. Moreover, even among the admissible past positions, their relevance for predicting the current stress is highly non-uniform. Positions near loading reversals, yield transitions, or damage onset typically carry far more information than positions within purely elastic response. To address both points, we incorporate a multi-head self-attention module \cite{Vaswani2017} with a strict causal mask, which adaptively reweights the temporal axis while preventing any position from attending future values, as shown in Figure~\ref{fig:FNO_SIREN_ATTENTION} and \ref{fig:atten}.

For an input sequence $\mathbf{X} \in \mathbb{R}^{N \times d}$, where $N$ is the sequence length (the number of discretized time steps along the loading trajectory) and $d$ is the feature dimension at that stage of the network, the attention block computes queries, keys, and values
\begin{equation}
\mathbf{Q} = \mathbf{X} W_Q, \qquad \mathbf{K} = \mathbf{X} W_K, \qquad \mathbf{V} = \mathbf{X} W_V,
\end{equation}
where $W_Q, W_K \in \mathbb{R}^{d \times d_k}$ and $W_V \in \mathbb{R}^{d \times d_v}$ are learnable projection matrices, $d_k$ is the dimension of the query and key vectors, and $d_v$ is the dimension of the value vectors (and hence of each output row of the attention block). The block then returns
\begin{equation}
\mathrm{Attn}(\mathbf{X}) = \mathrm{softmax}\!\left( \frac{\mathbf{Q} \mathbf{K}^\top}{\sqrt{d_k}} + \mathbf{M} \right) \mathbf{V},
\end{equation}
where $\mathrm{softmax}$ denotes the function that maps a real-valued vector $\mathbf{z} \in \mathbb{R}^n$ to a probability distribution via $\mathrm{softmax}(\mathbf{z})_j = \exp(z_j) / \sum_{k=1}^n \exp(z_k)$, and is applied row-wise here, turning each row of $\frac{\mathbf{Q}\mathbf{K}^\top}{\sqrt{d_k}} + \mathbf{M}$ into a probability distribution over time steps, so that each output position is a weighted average of the value vectors in $\mathbf{V}$. The factor $\sqrt{d_k}$ rescales the query-key dot products, which otherwise grow in magnitude with $d_k$ and push the softmax into a regime with vanishingly small gradients. The causal mask $\mathbf{M} \in \mathbb{R}^{N \times N}$ has entries
\begin{equation}
M_{ij} = \begin{cases} 0, & j \leq i, \\ -\infty, & j > i, \end{cases}
\end{equation}

The masked entries vanish under the softmax and the attention output at position $i$ is a convex combination of value vectors at positions $j \leq i$ only. Multiple heads operate in parallel and their outputs are concatenated through a linear projection. The mask is what enforces causality at the architectural level. The causal self-attention layer operates on hidden feature representations of the strain history, not on stress. It is combined with the pointwise SIREN layers (which act independently at each position) and with the spectral convolution, which is applied after the masked attention (Figure \ref{fig:FNO_SIREN_ATTENTION}).

Each operator block in the proposed architecture combines causal attention, spectral convolution, and a SIREN feature transform with residual connections:
\begin{equation}
\mathbf{v} \;\leftarrow\; \mathbf{v} + \mathrm{Attn}\big(\mathrm{LN}(\mathbf{v})\big), \qquad \mathbf{v} \;\leftarrow\; \sin\!\big( \mathcal{K}_\ell \mathbf{v} + W_\ell \mathbf{v} \big),
\end{equation}

where $\mathrm{LN}$ is layer normalization, $\mathcal{K}_\ell$ is the spectral convolution , $W_\ell$ is a pointwise SIREN linear map, and the sine activation provides the final nonlinearity of the block. Stacking $L$ such blocks between a SIREN lifting operator $\mathcal{P}$ and a SIREN-plus-linear projection $\mathcal{Q}$ produces the full FNO-SIREN operator $\mathcal{G}$. To reduce the risk of overfitting, meaning the model memorizing the training data instead of learning the underlying operator, we apply dropout during training. Dropout works by randomly setting a fraction of the intermediate values in the network to zero at each training step, with the specific values chosen anew every time. Dropout is applied inside each attention block and after each spectral activation for regularization.

A minor implementation detail worth noting is that strain-to-stress sequences are zero-anchored: the first entry of every training trajectory satisfies $\boldsymbol{\varepsilon}_0 = \mathbf{0}$ and $\boldsymbol{\sigma}_0 = \mathbf{0}$, so the output of the network is post-processed by subtracting ${\boldsymbol{\sigma}}_0$ from every position. This ensures the zero-stress boundary condition is satisfied exactly regardless of any residual bias in the projection head.

\subsubsection{Training protocol}
\textbf{Loss:} The network is trained with the mean squared error over the full predicted stress trajectory:
\begin{equation}
\mathcal{L}_{\mathrm{MSE}} = \frac{1}{N_{\mathrm{tr}}} \sum_{j=1}^{N_{\mathrm{tr}}} \frac{1}{N_j} \sum_{n=0}^{N_j-1} \big( \hat{\boldsymbol{\sigma}}_n^{(j)} - \boldsymbol{\sigma}_n^{(j)} \big)^2,
\end{equation}
where $N_{\mathrm{tr}}$ denotes the number of training trajectories (paths) in the dataset, $N_j$ is the number of time steps in the $j$-th trajectory, and $\hat{\boldsymbol{\sigma}}^{(j)} = \mathcal{G}(\boldsymbol{\varepsilon}^{(j)})$ is the prediction for the $j$-th training trajectory, obtained by mapping the entire strain trajectory $\boldsymbol{\varepsilon}^{(j)}$ to the entire predicted stress trajectory $\hat{\boldsymbol{\sigma}}^{(j)}$ in a single forward pass.

Note that no autoregressive unrolling is needed, i.e., the model does not generate the trajectory step-by-step by recursively feeding back its own previous predictions. Strain and stress components are standardized to zero mean and unit variance using the training-set statistics, and predictions are denormalized before evaluation.

\textbf{Optimization:} All models are trained with AdamW \cite{loshchilov2018decoupled} with the learning rate, batch size, and weight decay determined by hyperparameter search. Early stopping with a patience parameter is used to avoid overfitting during training. The exact optimized hyperparameters used in our final training phase are reported.
Table~\ref{tab:hyperparameters} reports the tuned configurations for the two one-dimensional material models and the two-dimensional model.


\begin{table}[H]
    \centering
    \begin{tabularx}{\textwidth}{|>{\raggedright\arraybackslash}X|>{\raggedright\arraybackslash}X|>{\raggedright\arraybackslash}X|>{\raggedright\arraybackslash}X|}
        \hline
        \textbf{Hyperparameter} & \textbf{1D Elastoplastic model (see Section \ref{subsec: 1-d elasto})} & \textbf{1D Coupled Damage-Plasticity (see Section \ref{subsec:1d_damage})} & \textbf{2D Elastoplastic (see Section \ref{subsec:2d_elasto})} \\ \hline
        Channel width  & 96 & 96 & 48 \\ \hline
        Fourier modes per SpectralConv  & 16 & 4 & 12 \\ \hline
        Number of layers  & 6 & 4 & 5 \\ \hline
        SIREN activation frequency scaling ($\omega_0$) & 20.75 & 19.95 & 9.37 \\ \hline
        Dropout rate  & 0.11 & 0.03 & 0.0000825 \\ \hline
        Number of attention heads  & 4 & 4 & 4 \\ \hline
        Optimizer & AdamW  & AdamW & AdamW \\ \hline
        Learning rate  & $2.03 \times 10^{-4}$ & $3.70 \times 10^{-4}$ & $8.25 \times 10^{-4}$ \\ \hline
        Weight decay  & $2.93 \times 10^{-4}$ & $5.74 \times 10^{-5}$ & $4.96 \times 10^{-4}$ \\ \hline
        Batch size  & 128 & 32 & 64 \\ \hline
        Loss function & Mean Squared Error & Mean Squared Error & Mean Squared Error \\ \hline
        Algorithm & Optuna (TPE) & Optuna (TPE) & Optuna (TPE) \\ \hline
        Total tuning trials  & 100  & 100  & 100 \\ \hline
        Maximum epochs / Patience & 10,000 / 2,000 & 10,000 / 2,000 & 10,000 / 3,000 \\ \hline
    \end{tabularx}
    \caption{Optimized hyperparameters for the 1D and 2D material models described in Section 2.1-2.3.  }
    \label{tab:hyperparameters}
\end{table}

\textbf{Hyperparameter selection:} The architecture exposes several hyperparameters — the number of retained Fourier modes $M$, the latent channel width $w$, the depth $L$ of operator blocks, the SIREN frequency $\omega_0$, the number of attention heads $H$, the dropout rate, and the optimizer settings (learning rate, batch size, weight decay). These are tuned jointly via Bayesian optimization with the Tree-structured Parzen Estimator algorithm \cite{Bergstra2011} as implemented in Optuna \cite{Akiba2019}, with median pruning to terminate unpromising trials early. Width and number of heads are drawn jointly from a predefined set of compatible pairs rather than independently. Each configuration is evaluated by its best validation loss over a reduced training budget; the best configuration identified by the search is then retrained at full budget and used for all reported results.

\textbf{Inference:} The full strain trajectory of length $n$ is presented to the model as a single tensor of shape $(B, N, C)$, where $B$ is the batch size, $N$ is the number of time steps, and $C$ is the number of strain components ($C=1$ for the 1D case, $C=3$ for 2D). The operator maps this to a stress tensor of identical shape $(B, N, C)$ in a single forward pass. There is no sliding window, no recurrent loop, and no prediction-feedback. The trajectory length $n$ at inference need not equal the training trajectory length. The operator can be evaluated on the same physical loading path sampled at any resolution, a property that is demonstrated in the discretization-sensitivity study.

\subsection{ Autoregressive Feed Forward Neural Networks with explicit history windows }

An FFNN realizes the discrete mapping by retaining only a finite recent window of past strain–stress observations and predicting the next stress step from this window together with the upcoming strain increment. Following \citet{Noor2026} and related studies, the network defines a parametric map

\begin{equation}
{{\sigma}}_{n+1} = \mathrm{FFNN}_\theta\!\big( {\varepsilon}_{n-W+1}, {\sigma}_{n-W+1}, \ldots, {\varepsilon}_n, {\sigma}_n, {\varepsilon}_{n+1} \big),
\end{equation}

where $W$ is the window length and $\theta$ collects all network parameters. The choice of $W$ is a hyperparameter; in our numerical experiments we report results for $W \in \{1, 5, 10\}$, denoted FFNN($W=1$), FFNN($W=5$), and FFNN($W=10$), to expose the sensitivity of FFNN performance to this choice.

\begin{figure} [H]
  \centering
  \includegraphics[width=0.90\linewidth]{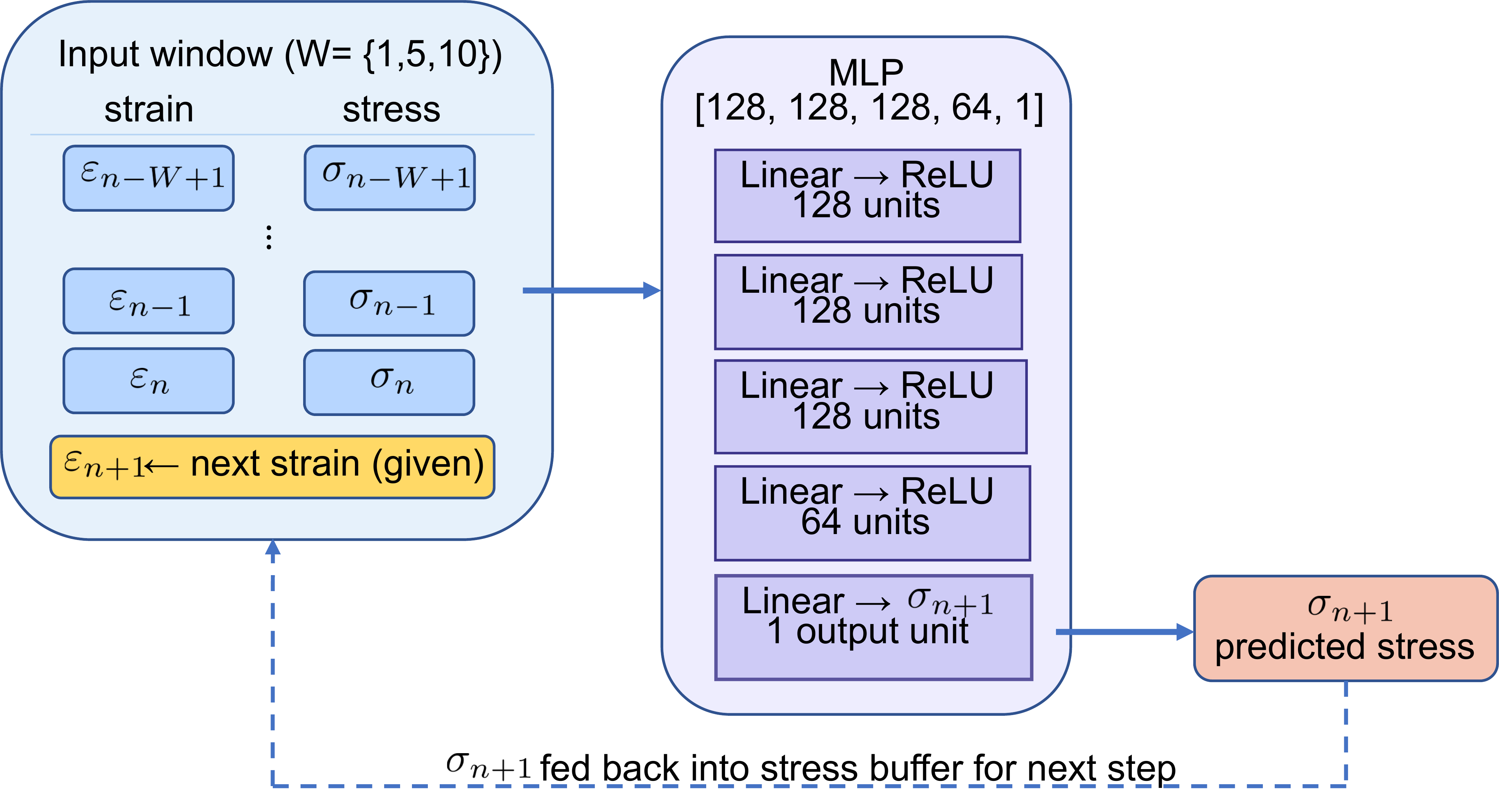}
  \caption{Autoregressive Inference: Feed Forward Neural Network Architecture}
  \label{fig:ffnn}
\end{figure}

Crucially, the FFNN operates autoregressively at inference. To produce a full stress trajectory of length $n$, the network is evaluated $n-W+1$ times in sequence, and after each evaluation the predicted stress ${{\sigma}}_{n+1}$ is fed back into the input window in place of the (unobserved) true stress. This introduces two well-known difficulties. First, errors accumulate along the trajectory because each step's prediction conditions every subsequent step as shown in Figure \ref{fig:ffnn}. Second, the input dimension scales linearly with $W$, making long-memory representations expensive and prone to overfitting; conversely, too small a window cannot capture the relevant material memory.

\subsubsection{Training protocol}
\textbf{Loss:} The FFNN is trained using the mean squared error over single-step predictions, utilizing overlapping historical windows extracted from the full trajectories:
\begin{equation}
\mathcal{L}_{\mathrm{MSE}} = \frac{1}{N_{\mathrm{pairs}}} \sum_{k=1}^{N_{\mathrm{pairs}}} \big( \hat{\boldsymbol{\sigma}}_{n+1}^{(k)} - \boldsymbol{\sigma}_{n+1}^{(k)} \big)^2,
\end{equation}
where $N_{pairs}$ are total numbers of discrete points in all trajectories/paths given for training, $\hat{\boldsymbol{\sigma}}_{n+1}^{(k)}(\theta) = f_\theta(\boldsymbol{\varepsilon}_{n-W+1:n}^{(k)}, \boldsymbol{\sigma}_{n-W+1:n}^{(k)}, \boldsymbol{\varepsilon}_{n+1}^{(k)})$ is the prediction for the next stress state given a sliding window of size $W$. Unlike the sequence-to-sequence FNO model, the FFNN relies on historical stresses during training. The model is trained exclusively on fully populated history windows starting from time step $W$.

\textbf{Optimization:} The models are optimized using AdamW (\citet{loshchilov2018decoupled}) with a fixed learning rate of $10^{-3}$, a weight decay of $10^{-4}$, and a batch size of 512. Training proceeds for a maximum of 10,000 epochs, utilizing early stopping with a patience of 500 epochs evaluated on the validation set's mean squared error to prevent overfitting.

\textbf{Hyperparameter selection:} In contrast to the Bayesian optimization employed for the operator learning approach, the FFNN utilizes a fixed multi-layer perceptron architecture. For example, the hyperparameters chosen for 1D elastoplastic material of Section \ref{subsec: 1-d elasto} were four hidden layers with sizes $\{128, 128, 128, 64\}$ with ReLU activations. The primary structural hyperparameter investigated is the sliding window size $W$, which dictates the temporal memory capacity of the model. Specifically, models with $W \in \{1, 5, 10\}$ are evaluated independently to quantify the effect of historical context length on path-dependent predictive accuracy. Consequently, the input feature dimension is $2W+1$, comprising $2W$ past strain and stress values, plus the current strain for which the corresponding stress is being evaluated.

\textbf{Inference:} During deployment, the FFNN operates in a strictly autoregressive manner, a fundamental departure from the non-autoregressive FNO. The model must feed its own past predictions back into the input buffer to step forward in time. The autoregressive unrolling takes the form
\begin{equation}
\hat{\boldsymbol{\sigma}}_{n+1} = f_\theta(\boldsymbol{\varepsilon}_{n-W+1:n}, \hat{\boldsymbol{\sigma}}_{n-W+1:n}, \boldsymbol{\varepsilon}_{n+1}),
\end{equation}
where $\hat{\boldsymbol{\sigma}}$ denotes the model's predictions. Both the sequence strain and sequence stress rolling buffers are initialized to zero to properly enforce the physical initial condition of the material prior to loading.

\subsection{Recurrent neural networks with hidden-state memory}
\begin{figure}[H] 
  \centering
\includegraphics[width=0.65\linewidth]{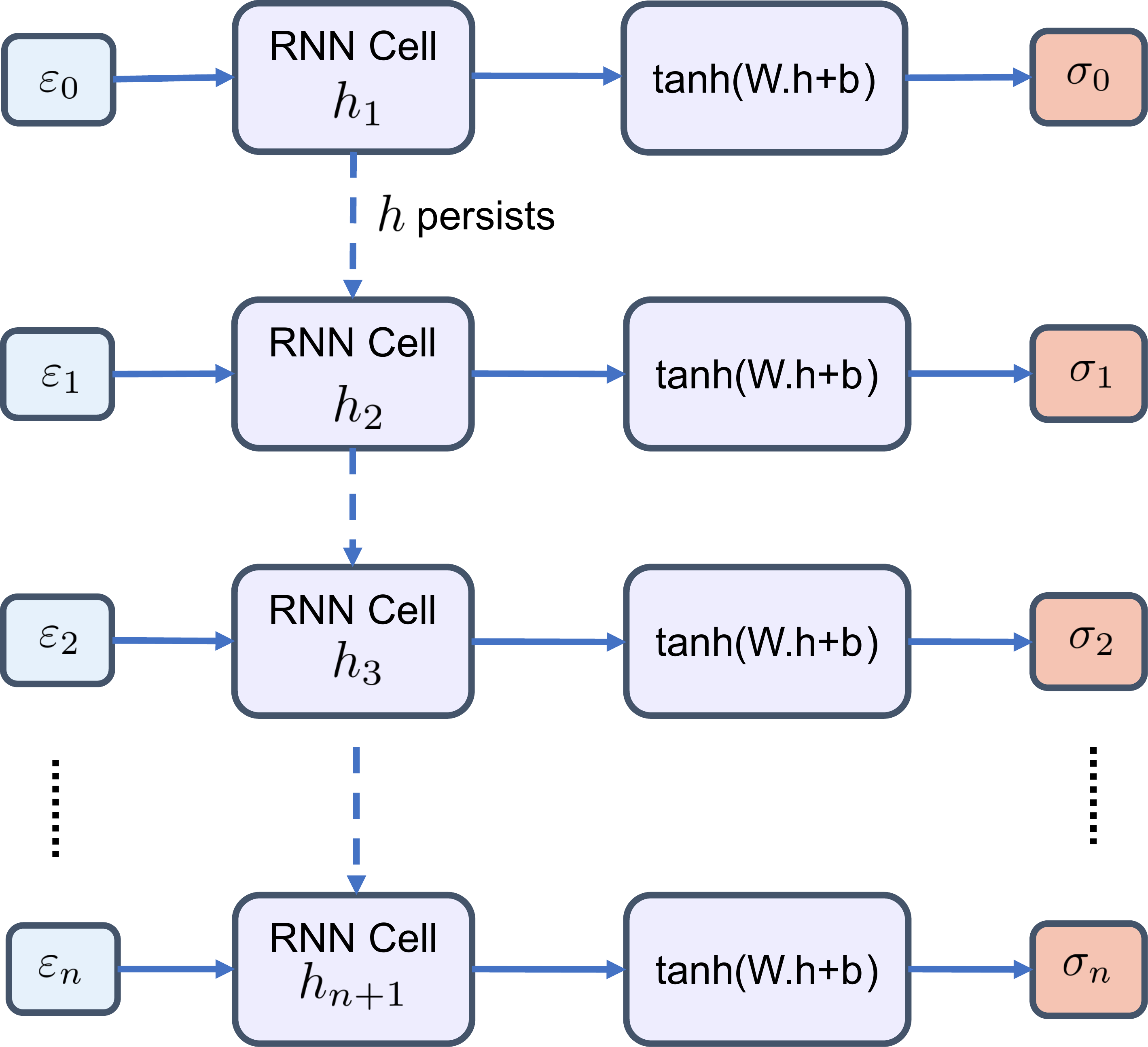}
  \caption{RNN architecture variant with GRU cell}
  \label{fig:RNN}
\end{figure}

RNN and its gated variants, in particular GRU and LSTM (Long Short-Term Memory) cells, provide an alternative representation in which the loading memory is encoded in an evolving hidden state $\mathbf{{h}_n}$ rather than in an explicit input window. Conceptually the hidden state plays the role of a learned, nominal counterpart to the physical internal state variables. It carries the loading memory forward in time without requiring the user to specify what that memory should contain as shown in Figure \ref{fig:RNN}. RNN-based surrogates have been applied to a broad range of inelastic problems including plasticity (\citet{Mozaffar2019,Gorji2020}). Like the FFNN, the RNN processes the loading path sequentially. A trajectory of length $n$ requires $N$ sequential network evaluations.

The principal structural limitation of RNN-based surrogates in this context is \textit{discretization dependence}. Because the hidden state is updated through a per-step nonlinear recurrence, the trajectory of $\mathbf{h}_n$ depends on the training resolution of a given loading path. Refining the temporal sampling, even when the underlying continuous strain history is unchanged, produces a different sequence of hidden states and hence a different stress prediction. This behavior violates the self-consistency expected of a rate-independent constitutive law.

\subsubsection{Training protocol}
\textbf{Loss:} The RNN is trained using the mean squared error evaluated over the entirety of the predicted stress trajectory:
\begin{equation}
\mathcal{L}_{\mathrm{MSE}} = \frac{1}{N_{\mathrm{tr}}} \sum_{j=1}^{N_{\mathrm{tr}}} \frac{1}{N_j} \sum_{n=0}^{N_j-1} \big( \hat{\boldsymbol{\sigma}}_n^{(j)} - \boldsymbol{\sigma}_n^{(j)} \big)^2,
\end{equation}
where $\hat{\boldsymbol{\sigma}}_n^{(j)}$ is the stress predicted at timestep $n$. The model processes the strain history sequentially, taking only the single instantaneous strain tensor $\boldsymbol{\varepsilon}_n$ as input at each step. The path-dependency is captured entirely by the internal hidden state $\boldsymbol{h}_n$, which evolves continuously via the Gated Recurrent Unit (GRU) operations and is never reset between timesteps.

\textbf{Optimization:} The network is trained using the AdamW optimizer \cite{loshchilov2018decoupled} with a fixed learning rate of $10^{-3}$, a weight decay of $10^{-4}$, and a batch size of 32. The training process runs for a maximum of 500 epochs, utilizing early stopping with a patience of 200 epochs monitored on the validation set's mean squared error to prevent overfitting.

\textbf{Hyperparameter selection:} Instead of conducting a Bayesian hyperparameter search, this baseline adheres to the exact architectural specifications proposed by \citet{Gorji2020}. The GRU model is constructed with a fixed configuration: $N_{\mathrm{GRU}} = 2$ stacked GRU layers, each containing $N_{\mathrm{NPU}} = 100$ hidden neurons. The final hidden state is passed through a Fully Connected Neural Network (FCNN) featuring $N_{\mathrm{HL}} = 1$ hidden layer with $N_{\mathrm{NPL}} = 100$ neurons to map the latent memory representation $\boldsymbol{h}_n$ to the instantaneous stress state $\boldsymbol{\sigma}_n$.

\textbf{Inference:} The sequence of strains is processed in a single forward sequential pass. At each discrete timestep $n$, the single strain value $\boldsymbol{\varepsilon}_n$ updates the persistent hidden state $\boldsymbol{h}_{n-1} \to \boldsymbol{h}_n$, and the shared FCNN maps this updated state to $\hat{\boldsymbol{\sigma}}_n$. Unlike the FFNN approach, there is no sliding window of historical data to manage, and unlike typical sequence generation models, there is no feedback loop of prior predictions into the input stream. The temporal memory is strictly encapsulated within the continuous evolution of the GRU hidden state.

\section{Results}

This section evaluates the proposed FNO-SIREN with causal attention and compares it against two families of baselines: feedforward networks with sliding history windows of length $W \in \{1, 5, 10\}$, and an RNN-GRU surrogate. The evaluation is done for three material models: a one-dimensional rate-independent elastoplastic model with nonlinear isotropic hardening, a one-dimensional rate-independent coupled damage-plasticity model and a two-dimensional rate-independent elastoplastic model with a nonlinear isotropic hardening (see Sections 2.1-2.3, respectively). 
We probe the central methodological claim of this work, discretization invariance, by evaluating the trained models on physically identical loading paths sampled at temporal resolutions ranging from $N = 50$ (the training resolution) up to $N = 1000$.

\subsection{One-dimensional elastoplastic material model}
\subsubsection{Datasets}
\label{subsubsec: Datasets_1d}
\begin{figure}[H]
  \centering
  \begin{subfigure}[b]{0.48\textwidth}
    \centering
    \includegraphics{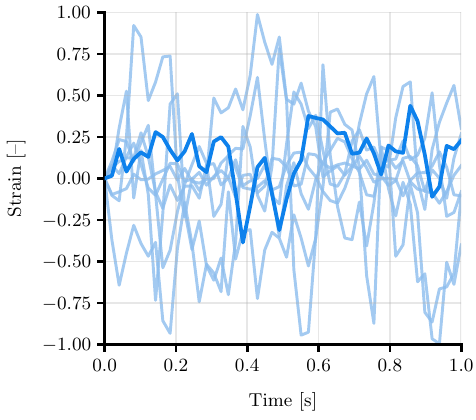}
    \label{fig:train_strain_1d_plasti}
  \end{subfigure}
  \hspace{0.03\textwidth}%
  \begin{subfigure}[b]{0.48\textwidth}
    \centering
    \includegraphics{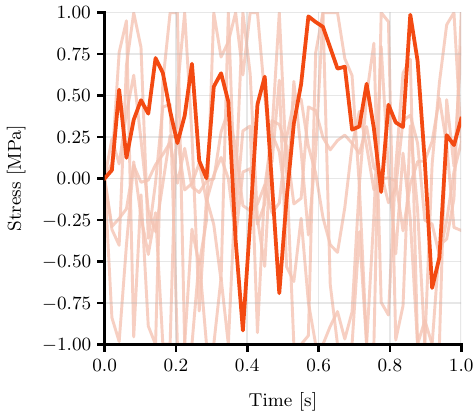}
    \label{fig:train_stress_1d_plasti}
  \end{subfigure}
  \caption{Training distribution of strain paths (blue, left) and corresponding stress responses (orange, right). The bold solid curves highlight representative strain–stress pairs, while the faint curves are some remaining trajectories in the dataset.  } 
  \label{fig:training distribution for 1d plasti} 
\end{figure}

We generate $N_{\mathrm{tr}} =9000$ Gaussian-Process (GP) loading paths for training and another $N_{\mathrm{val}} = 1000$ for validation, all uniformly sampled at $N = 50$ steps over a fixed time interval. Each path is drawn from a zero-mean GP with squared-exponential covariance whose length scale $\ell$ is sampled uniformly from $[0.005, 0.05]$, shifted so that $\varepsilon_0 = 0$, and rescaled so that the peak strain is drawn uniformly from $[-1.0, 1.0]$. This produces a wide distribution of smooth, randomly oscillating loading paths that exercise repeated yielding, unloading, and reloading and the corresponding stress paths were also generated based on the material model defined in Section \ref{subsec: 1-d elasto}. In Figure \ref{fig:training distribution for 1d plasti}, some training strain paths together with the corresponding stress paths are shown.

\begin{figure}[H]
  \centering
  
  \begin{subfigure}[b]{0.49\linewidth}
    \centering
    \includegraphics{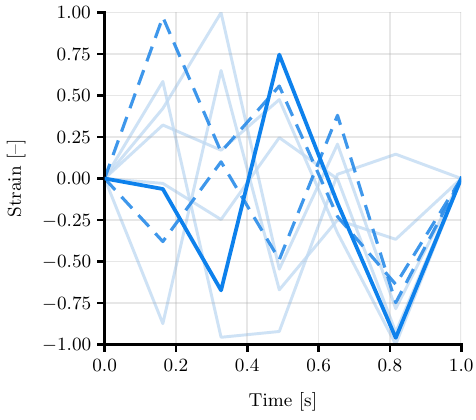}
    \label{fig:testing_distri_zig}
  \end{subfigure}
  \hfill
  \begin{subfigure}[b]{0.49\linewidth}
    \centering
    \includegraphics{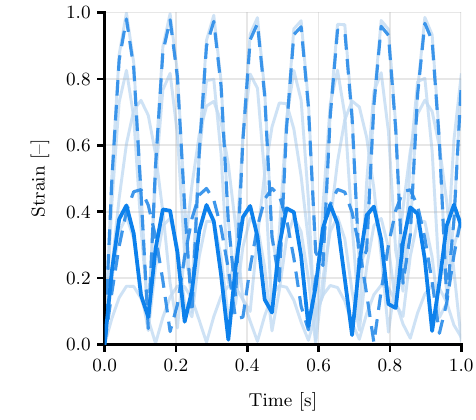}
    \label{fig:testing_distri_sin}
  \end{subfigure}

  \vspace{0.8cm} 

  \begin{subfigure}[b]{0.49\linewidth}
    \centering
    \includegraphics{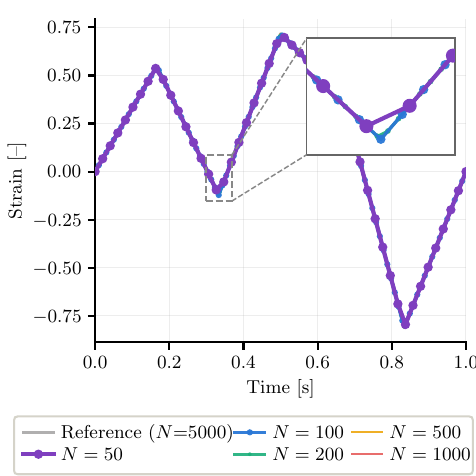} 
    \label{fig:testing_distri_zig_disc}
  \end{subfigure}
  \hfill
  \begin{subfigure}[b]{0.48\linewidth}
    \centering
    \includegraphics{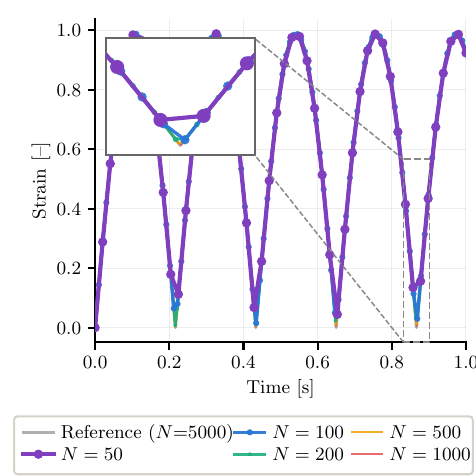} 
    \label{fig:testing_distri_sin_dis}
  \end{subfigure}

  \caption{Testing distribution and discretization study. Top row shows sample strain paths from the two testing distributions, zig-zag (left) and sinusoidal (right), with bold and dashed curves marking representative examples among the faint background paths. Bottom row shows the same paths resampled at different resolutions (N = 50 to N = 1000) against a high resolution reference (N = 5000).} 
  \label{fig:testing_distri} 
\end{figure}

Three families of loading test paths are used to evaluate generalization, each containing 100 trajectories at $N = 50$:

\begin{itemize}
    \item \textbf{Gaussian-Process paths:} Freshly drawn GP paths, in-distribution with the training set, used as a sanity check.
    \item \textbf{Zig-zag paths:} piecewise-linear paths defined by seven equally spaced key points whose interior values are sampled uniformly from $[-1, 1]$ as shown in Figure \ref{fig:testing_distri}.
    \item \textbf{Sinusoidal paths:} $\varepsilon(t) = a \,|\sin(2\pi f\, t)|$ with frequency $f$ uniform on $[1, 2.5]$ and amplitude $a$ uniform on $[0.1, 1.0]$. These paths contain multi-cycle, oscillatory loading--unloading sequences as shown in Figure \ref{fig:testing_distri}.
\end{itemize}

In Figure \ref{fig:testing_distri}, the images in the second row show testing paths used to evaluate the models when a single strain loading is sampled at different resolutions, in order to assess discretization invariance. Further details on how these plots were used for the discretization study are provided in Section \ref{subsec: Discretization_study_1d_elasto}.

\subsubsection{Representative predictions}
\label{sec:representative predictions 1D elasto}
To illustrate the qualitative behavior of the three model families before 
presenting aggregate statistics, we examine representative stress-strain 
hysteresis loops under two evaluation conditions. 

\begin{figure}[H]
  \centering
  
  \begin{subfigure}[b]{0.49\linewidth}
    \centering \includegraphics{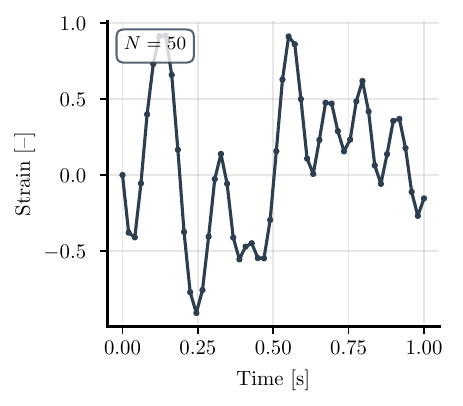}
    \subcaption{Strain loading path at training resolution}
    \label{fig:test_in_temp50}
  \end{subfigure}
  \hfill
  \begin{subfigure}[b]{0.49\linewidth}
    \centering
    \includegraphics{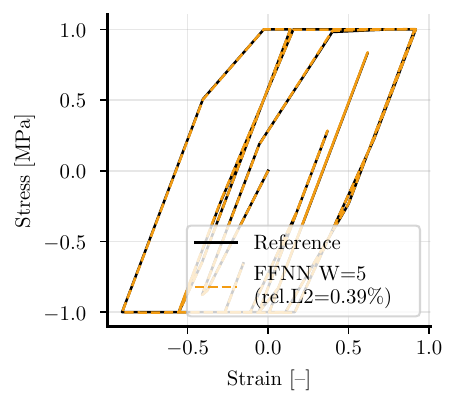} 
    \subcaption{FFNN(W = 5)}
    \label{fig:test_in_ffnn}
  \end{subfigure}
  \vfill
  \begin{subfigure}[b]{0.49\linewidth}
    \centering
    \includegraphics{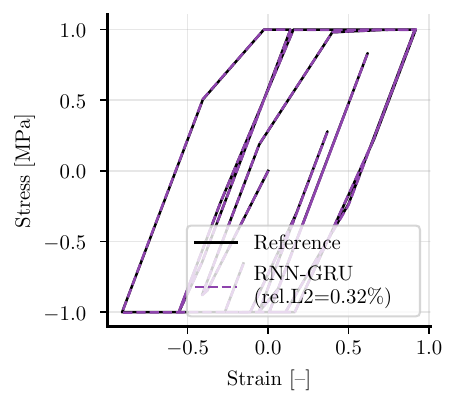} 
    \subcaption{RNN-GRU}
    \label{fig:test_in_RNN}
  \end{subfigure}
    \hfill
  \begin{subfigure}[b]{0.49\linewidth}
    \centering
    \includegraphics{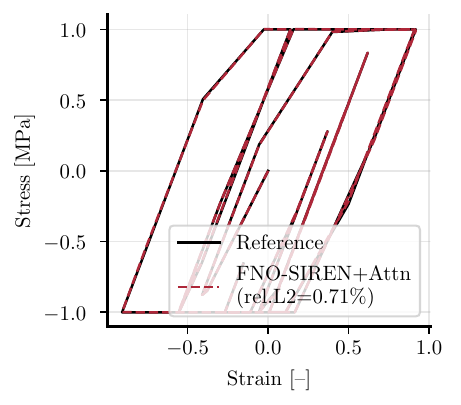} 
    \subcaption{FNO-SIREN+Attention}
    \label{fig:test_in_PO}
  \end{subfigure}

  \caption{In-distribution (Gaussian-Process) loading path at temporal resolution N = 50}
  \label{fig:in_distri_temp50}
\end{figure}

In each case we compare the FFNN with window size 5, the RNN-GRU surrogate, and the proposed 
FNO-SIREN with causal attention operator. 
The two conditions are 
(i)~a Gaussian-Process path evaluated at the training 
resolution $N=50$ as shown in Figure \ref{fig:test_in_temp50}, which serves as a sanity check; 
(ii) a complex, variable-amplitude cyclic loading path (evaluated at $N=1000$) as shown in Figure \ref{fig:test_out_temp500} designed to simultaneously test path-family generalization and resolution invariance.

\begin{figure}[H]
  \centering
  
  \begin{subfigure}[b]{0.49\linewidth}
    \centering
    \includegraphics{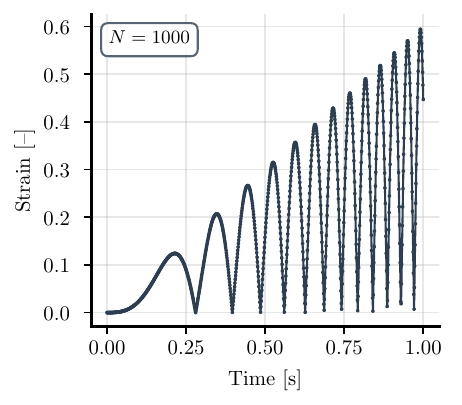}
    \subcaption{Strain loading path at temporal resolution N = 1000}
    \label{fig:test_out_temp500}
  \end{subfigure}
  \hfill
  \begin{subfigure}[b]{0.49\linewidth}
    \centering
    \includegraphics{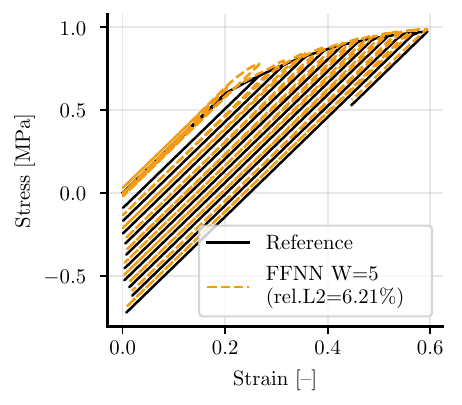} 
    \subcaption{FFNN(W = 5)}
    \label{fig:test_out_ffnn}
  \end{subfigure}
  \end{figure}
  \begin{figure}
  \ContinuedFloat
  \vfill
  \begin{subfigure}[b]{0.49\linewidth}
    \centering
    \includegraphics{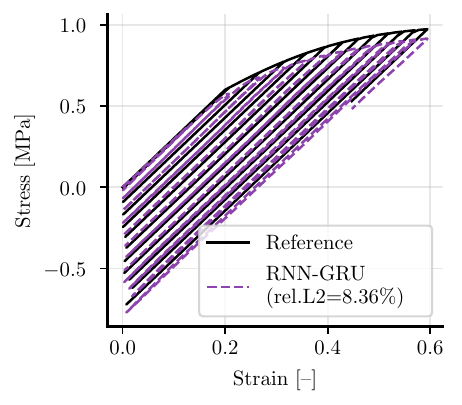} 
    \subcaption{RNN-GRU}
    \label{fig:test_out_RNN}
  \end{subfigure}
    \hfill
  \begin{subfigure}[b]{0.49\linewidth}
    \centering
    \includegraphics{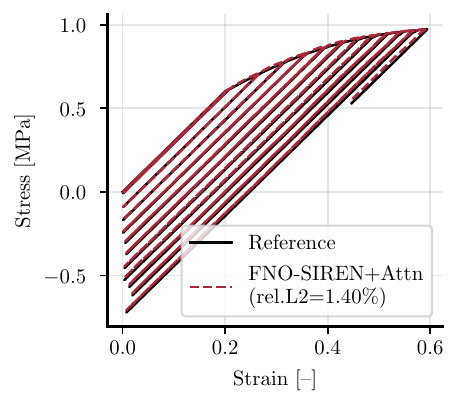} 
    \subcaption{FNO-SIREN+Attention}
    \label{fig:test_out_PO}
  \end{subfigure}

  \caption{Complex loading path at temporal resolution N = 1000 }
  \label{fig:out_distri_temp500}
\end{figure}

Figures \ref{fig:test_in_ffnn}, \ref{fig:test_in_RNN}, and \ref{fig:test_in_PO} show that, at the training resolution, all models predict quite accurately on paths similar to the Gaussian training paths. Reference solutions are obtained from the material model calculated numerically. Model accuracy is evaluated using the relative L2 error, defined in the formula below. 

\begin{equation}
    e_{L_2} = \frac{\left\lVert \hat{y} - y \right\rVert_2}{\left\lVert y \right\rVert_2} = \sqrt{\frac{\sum_{i=1}^{N} \left( \hat{y}_i - y_i \right)^2}{\sum_{i=1}^{N} y_i^2}}
    \label{eq:relative_l2_error}
\end{equation}

where $\hat{y}_i$ is the predicted stress at point $i$, $y_i$ is the corresponding reference value from the numerical material model, and $N$ is the number of points along the loading path. All models achieve a relative error below 1\% with respect to the reference solution.

However, when evaluated at a higher resolution with more complex loading paths, as shown in Figures \ref{fig:test_out_ffnn}, \ref{fig:test_out_RNN}, and \ref{fig:test_out_PO}, the FFNN and RNN-GRU models exhibit an abrupt increase in error, ranging from 6–9\%, while the proposed operator architecture, FNO-SIREN+Attention, maintains a substantially lower error of 1.4\%. To further investigate this behavior, we conducted a statistical study to assess discretization invariance. 

\subsubsection{Discretization Study}
\label{subsec: Discretization_study_1d_elasto}
\begin{figure}[H]
  \centering
  
  \begin{subfigure}[b]{0.49\linewidth}
    \centering
    \includegraphics{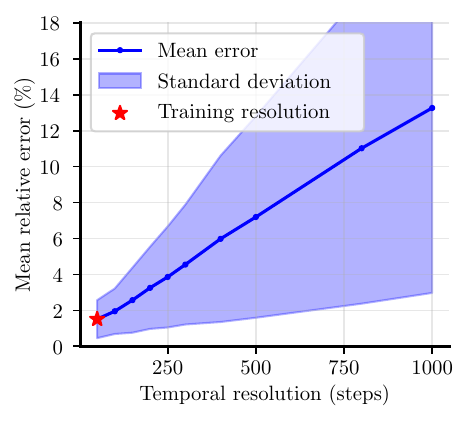}
    \subcaption{Feed Forward Neural Network with window size 1}
    \label{fig:test_1}
  \end{subfigure}
  \hfill
  \begin{subfigure}[b]{0.49\linewidth}
    \centering
    \includegraphics{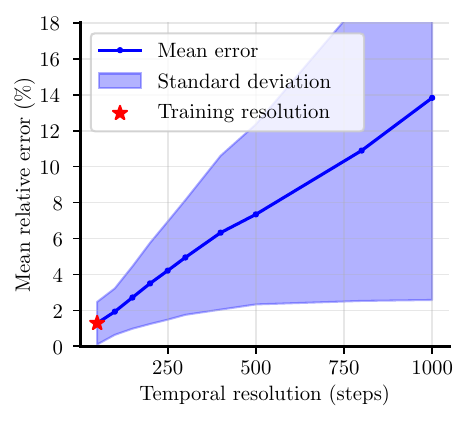}
    \subcaption{Feed Forward Neural Network with window size 5}
    \label{fig:test_2}
  \end{subfigure}

 \begin{subfigure}[b]{0.49\linewidth}
    \centering
    \includegraphics{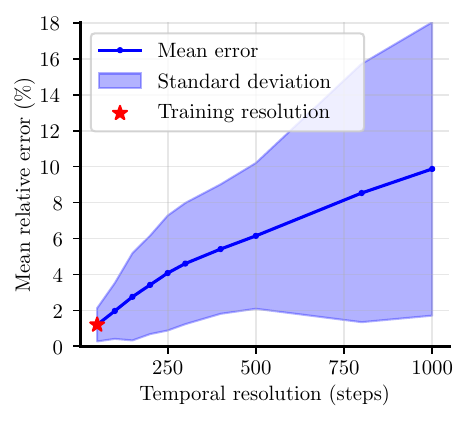} 
    \subcaption{Feed Forward Neural Network with window size 10}
    \label{fig:test_3}
  \end{subfigure}
  \hfill
  \begin{subfigure}[b]{0.49\linewidth}
    \centering
    \includegraphics{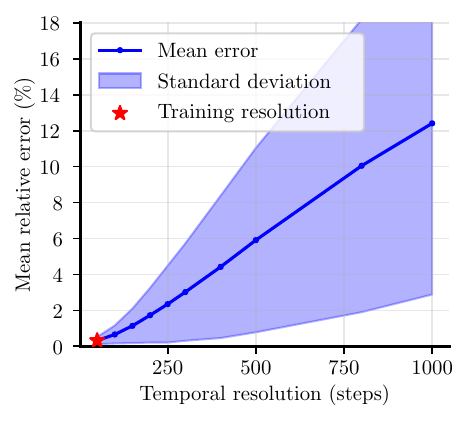} 
    \subcaption{RNN-GRU}
    \label{fig:test_4}
  \end{subfigure}

  \begin{subfigure}[b]{0.49\linewidth}
    \centering
    \includegraphics{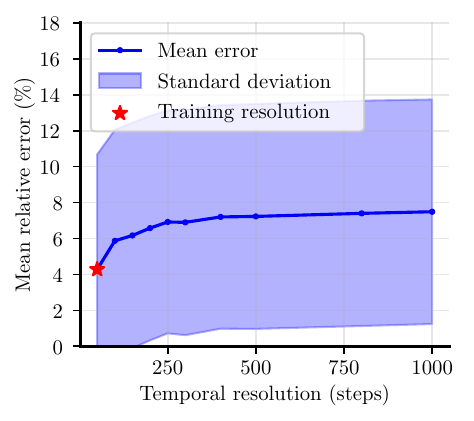} 
    \subcaption{FNO-SIREN}
    \label{fig:test_5}
  \end{subfigure}
  \hfill
  \begin{subfigure}[b]{0.49\linewidth}
    \centering
    \includegraphics{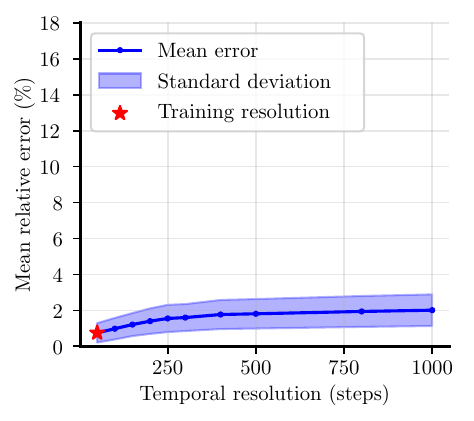} 
    \subcaption{FNO-SIREN+Attention}
    \label{fig:test_6}
  \end{subfigure}

  \caption{Discretization study for every individual model architecture}
  \label{fig:discretization_study_1d_elasto_results}
\end{figure}

\begin{figure}[H] 
  \centering
  \includegraphics{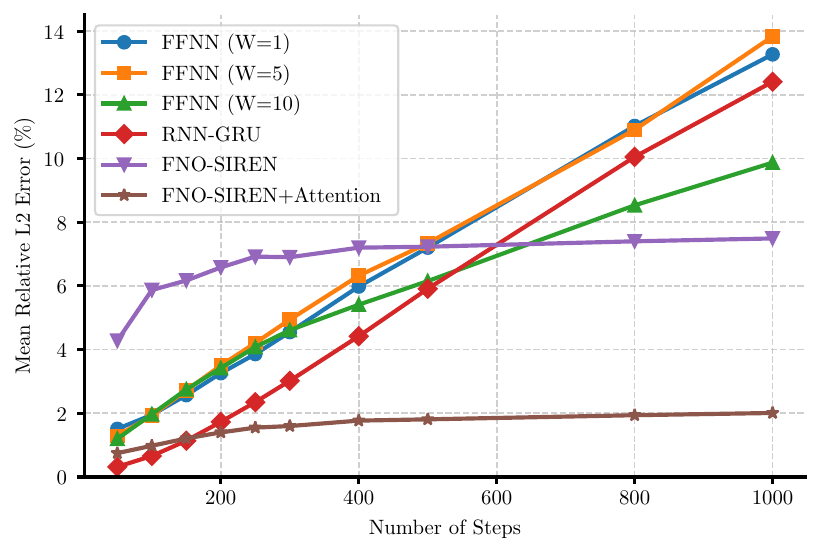}
  \caption{Discretization Study - Comparison Plot}
  \label{fig:discretization_study_for_all_elastoplastic_models}
\end{figure}

To probe discretization invariance, we evaluate six trained models on the canonical zig-zag test set (100 paths), with each path sampled at multiple temporal resolutions, as illustrated in Figure~\ref{fig:testing_distri}. Specifically, each of the 100 paths is resampled at $N$ $\in \{50$, $100$, $150$, $200$, $250$, $300$, $400$, $500$, $800$, $1000\}$  time steps, where $N$ denotes the number of steps. All models are trained at $N = 50$, and the underlying physical loading paths are identical across resolutions, differing only in temporal discretization. Per-model results across resolutions are reported in Figure~\ref{fig:discretization_study_1d_elasto_results}, and the six curves are combined in Figure~\ref{fig:discretization_study_for_all_elastoplastic_models} for direct comparison.

As shown in the graphs in Figure \ref{fig:discretization_study_1d_elasto_results}, the Feed Forward Neural Network with different window sizes and RNN-GRU baselines exhibit a clear trend, both the mean relative $L_2$ error and its standard deviation grow with sampling resolution. This behavior reflects a known structural limitation of RNN and FFNN based surrogates as the learned mapping is tied to the step size seen during training, and prediction errors accumulate over an increasing number of autoregressive evaluations as the path is refined. Mechanically, this contradicts the defining property of a rate-independent constitutive law that the stress response should depend on the loading path itself, not on the training resolution.

In contrast, the FNO-SIREN model mean error and standard deviation remain essentially constant (flat) throughout the entire resolution range. This is the expected consequence of operator learning. The spectral filters are indexed by physical wavenumbers rather than sample indices, so the same trained operator applies consistently to fine discretizations. Adding causal self-attention on top of the FNO-SIREN backbone further reduces the error at every resolution and significantly lowers the standard deviation compared to other model architectures, as shown in Figures~\ref{fig:discretization_study_1d_elasto_results} and~\ref{fig:discretization_study_for_all_elastoplastic_models}. This makes the model more robust while preserving discretization invariance. The attention-augmented model achieves a lower error at $N = 50$ and maintains essentially the same error at $N = 1000$, twenty times the training resolution. This is empirical evidence that the network has learned the underlying continuous strain-to-stress mapping rather than memorizing a step-size-specific approximation. We performed the same discretization study on 100 sinusoidal loading paths and observed the same overall trend.

\subsection{One-dimensional coupled damage-plasticity material model} 

\subsubsection{Datasets}
\begin{figure}[H]
  \centering
  \begin{subfigure}[b]{0.48\textwidth}
    \centering
    \includegraphics{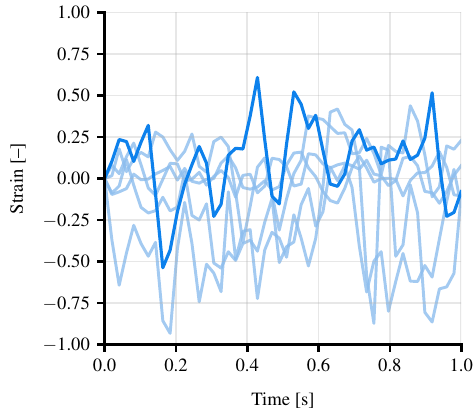}
    \label{fig:train_strain_1d_plasti}
  \end{subfigure}
  \hspace{0.03\textwidth}%
  \begin{subfigure}[b]{0.48\textwidth}
    \centering
    \includegraphics{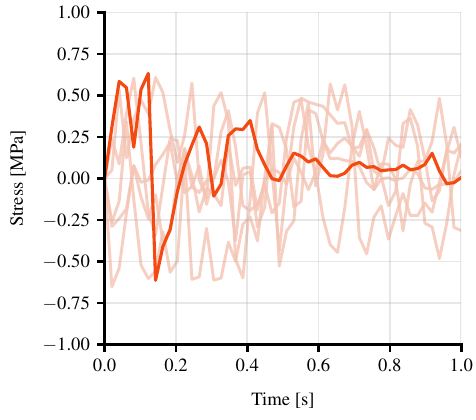}
    \label{fig:train_stress_1d_plasti}
  \end{subfigure}
  \caption{Training distribution: strain paths with corresponding stress paths  } 
  \label{fig:training distribution for 1d plasti damage-plasti} 
\end{figure}

The dataset construction follows exactly the protocol of Section \ref{subsubsec: Datasets_1d}. Training and validation paths are drawn from the same Gaussian-Process distribution at $N=50$ steps as shown in Figure \ref{fig:training distribution for 1d plasti damage-plasti}, with identical length-scale and amplitude ranges; the three test families (Gaussian-Process, zig-zag, and sinusoidal, 100 trajectories each) and the canonical zig-zag set used for the discretization study are generated with the same procedures. The reference stress responses are computed by a return-mapping algorithm using Newton iterations; all other dataset parameters are unchanged. This deliberate choice allows the comparison to isolate the effect of the underlying constitutive law from any variation in the loading-path statistics.

\subsubsection{Representative predictions}
\label{sec:representative predictions 1D ductile-damage}

\begin{figure}[H]
  \centering
  
  \begin{subfigure}[b]{0.49\linewidth}
    \centering
    \includegraphics{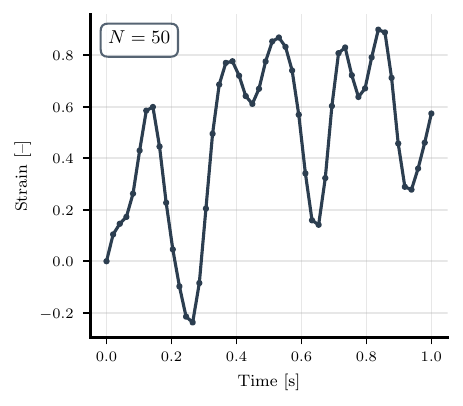}
    \subcaption{Strain loading path at training resolution}
    \label{fig:test_in_temp50_coupled}
  \end{subfigure}
  \hfill
  \begin{subfigure}[b]{0.49\linewidth}
    \centering
    \includegraphics{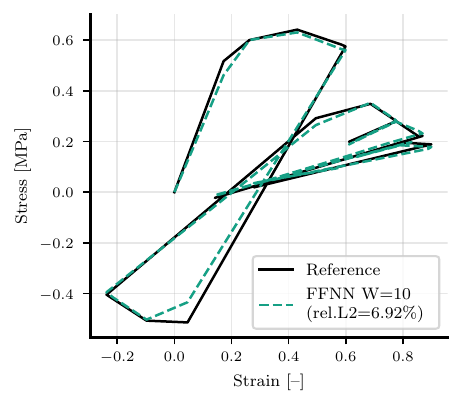} 
    \subcaption{FFNN(W = 10)}
    \label{fig:test_in_ffnn_coupled}
  \end{subfigure}

  \begin{subfigure}[b]{0.49\linewidth}
    \centering
    \includegraphics{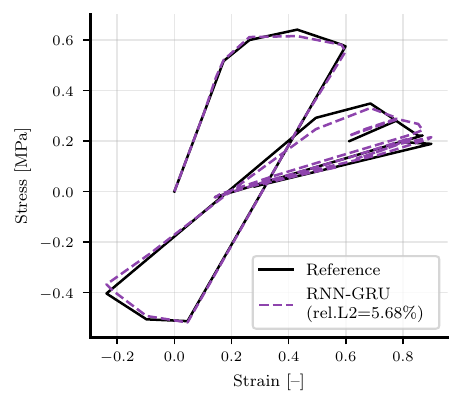} 
    \subcaption{RNN-GRU}
    \label{fig:test_in_RNN_coupled}
  \end{subfigure}
    \hfill
  \begin{subfigure}[b]{0.49\linewidth}
    \centering
    \includegraphics{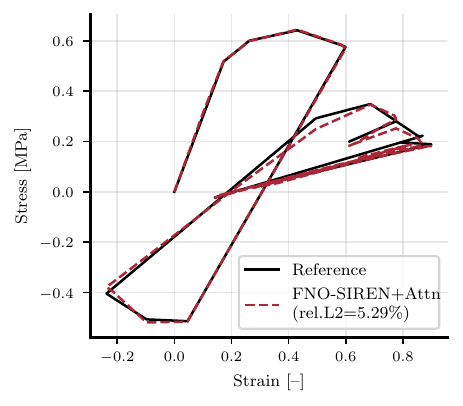} 
    \subcaption{FNO-SIREN+Attention}
    \label{fig:test_in_PO_coupled}
  \end{subfigure}

  \caption{In-distribution (Gaussian-Process) loading path at temporal resolution N = 50 }
  \label{fig:in_distri_temp50}
\end{figure}

To illustrate the qualitative behavior of the three model families for the damage-plasticity material model before 
presenting aggregate statistics, we examine representative stress-strain 
hysteresis loops under two progressively harder evaluation conditions. 
In each case we compare the FFNN with window size 10, the RNN-GRU surrogate, and the proposed 
FNO-SIREN with causal attention operator. The two evaluation conditions are 
(i)~a Gaussian-Process path evaluated at the training 
resolution $N=50$, which serves as a sanity check as shown in Figure \ref{fig:test_in_temp50_coupled}; 
(ii) a complex, variable-amplitude cyclic loading path (evaluated at $N=1000$) as shown in Figure \ref{fig:test_out_temp500_coupled} designed to simultaneously test path-family generalization and resolution invariance. All models are trained exclusively 
at $N=50$ on GP paths and receive no further training between conditions.

\begin{figure}[H]
  \centering
  
  \begin{subfigure}[b]{0.49\linewidth}
    \centering
    \includegraphics{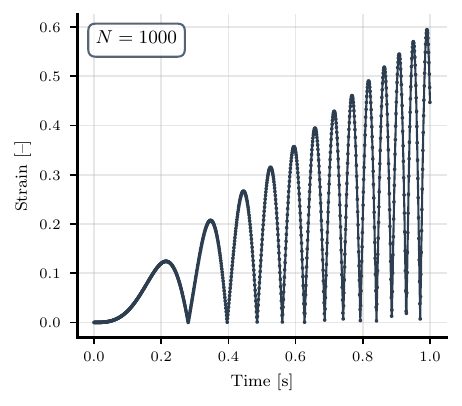}
    \subcaption{Strain loading path at temporal resolution N = 1000}
    \label{fig:test_out_temp500_coupled}
  \end{subfigure}
   \hfill
  \begin{subfigure}[b]{0.49\linewidth}
    \centering
    \includegraphics{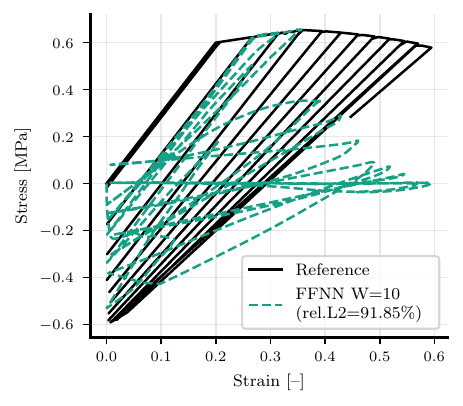} 
    \subcaption{FFNN(W = 10)}
    \label{fig:test_out_ffnn_coupled}
  \end{subfigure}
  \end{figure}
  
  \begin{figure}[H]
  \ContinuedFloat
  \begin{subfigure}[b]{0.49\linewidth}
    \centering
    \includegraphics{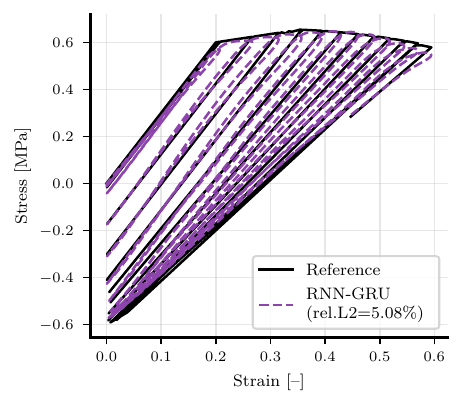} 
    \subcaption{RNN-GRU}
    \label{fig:test_out_RNN_coupled}
  \end{subfigure}
    \hfill
  \begin{subfigure}[b]{0.49\linewidth}
    \centering
    \includegraphics{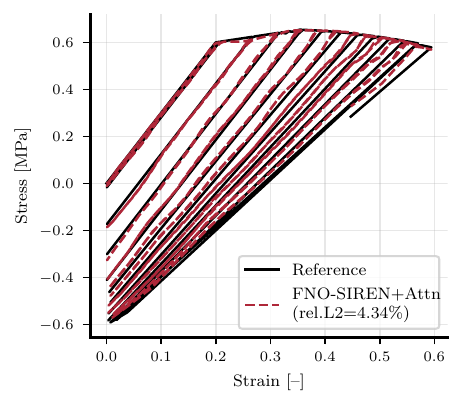} 
    \subcaption{FNO-SIREN+Attention}
    \label{fig:test_out_PO_coupled}
  \end{subfigure}

  \caption{Complex loading path at temporal resolution N = 1000 }
  \label{fig:out_distri_temp500_coupled}
\end{figure}

Figures \ref{fig:test_in_ffnn_coupled}, \ref{fig:test_in_RNN_coupled}, and \ref{fig:test_in_PO_coupled} show that, at the training resolution, all models predict quite similarly on paths similar to the Gaussian training paths. Reference solutions are obtained from the numerically calculated material model. Model accuracy is evaluated using the relative L2 error. All models had relative error with respect to the reference solution in the range of 5-7\%.

However, when evaluated at a higher resolution with more complex loading paths, as shown in Figures \ref{fig:test_out_ffnn_coupled}, \ref{fig:test_out_RNN_coupled}, and \ref{fig:test_out_PO_coupled}, the FFNN exhibits an abrupt increase in error of about 91\%, while the proposed operator architecture, FNO-SIREN+Attention, had the substantially lowest error of 4.34\%. To further investigate this behavior, we conducted a statistical study to assess discretization invariance. 

\subsubsection{Discretization Study}
\label{subsec: Discretization_study_1d_damage}
We repeat the discretization-invariance evaluation, with the surrogates now trained on the coupled damage-plastic model. The 100 canonical zig-zag test paths are evaluated at the same temporal resolutions $N \in \{50$, $100$, $150$, $200$, $250$, $300$, $400$, $500$, $800$, $1000\}$, and all six model architectures are again trained exclusively at $N=50$. Figure~\ref{fig:Discretization Study for all damage} reports the mean relative $L_2$ error as a function of $N$ for all six models. To make the comparison among the more competitive architectures visible, Figure~\ref{fig:Discretization Study for all damage} reproduces the same data with the three FFNN variants omitted, isolating the RNN-GRU baseline, the FNO-SIREN model, and the proposed FNO-SIREN with causal attention.

\begin{figure}[H]
\centering
 \begin{subfigure}[b]{0.8\linewidth}
  \centering
\includegraphics{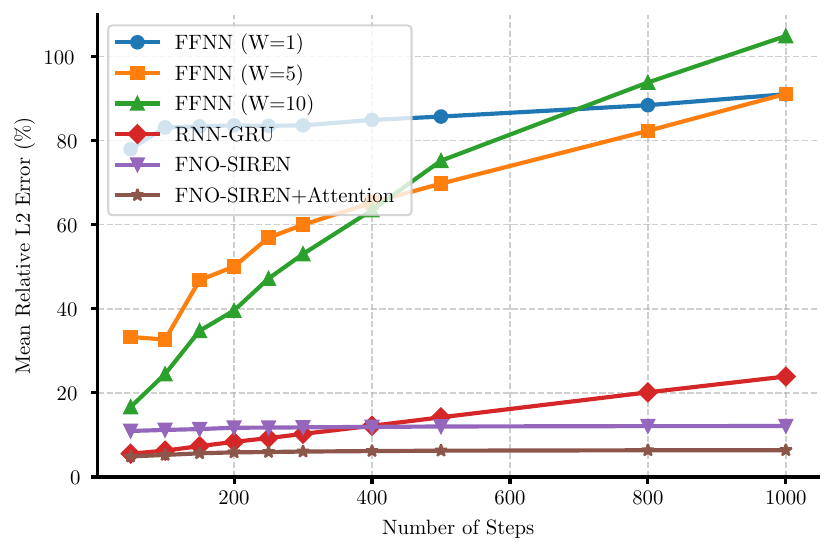}
  \subcaption{}
  \label{fig:Discretization Study for all damage a}
 \end{subfigure}
\end{figure}
\begin{figure}[H]
\continuedfloat
\centering
 \begin{subfigure}[b]{0.8\linewidth}
  \centering
\includegraphics{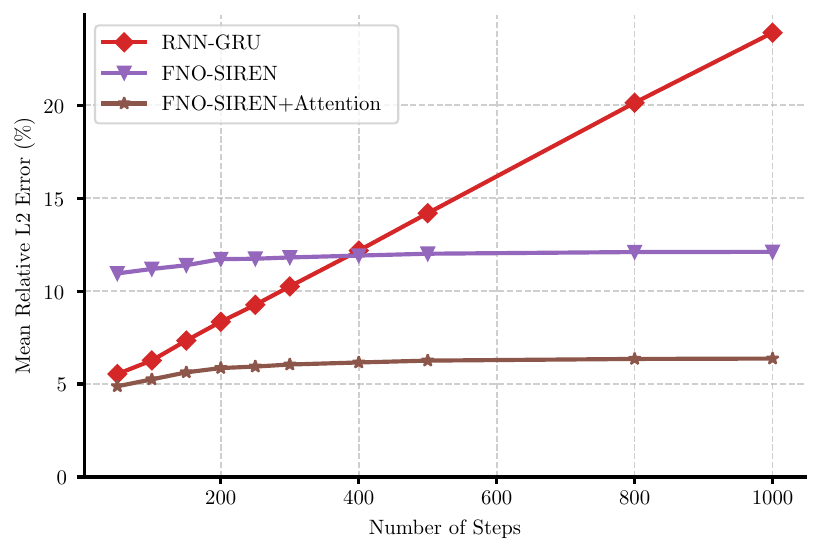}
  \subcaption{}
  \label{fig:Discretization Study for all damage b}
 \end{subfigure}
 \caption{Discretization study on the coupled plastic-damage model}
 \label{fig:Discretization Study for all damage}
\end{figure}

The qualitative picture is identical to that observed for pure elastoplasticity. FFNN accuracy degrades sharply as $N$ grows beyond the training resolution, with one exception: the FFNN with window size 1, whose error plateaus after a certain point as resolution increases. However, this error was already above 80\%, an anomalously high and somewhat erratic result that was not investigated further. The step-wise mapping is bound to the training-time discretization, and autoregressive rollout amplifies the mismatch as the path is refined. Adding damage to the constitutive law  makes the autoregressive drift even more visible as the network must simultaneously track plastic flow and stiffness degradation. The RNN-GRU surrogate  performs better than the FFNN family but exhibits a similar, milder upward drift in error  with $N$. The hidden-state recurrence, trained at $N=50$, is evaluated through finer-stepped trajectories at a pace that no longer matches the training distribution, and prediction errors accumulate accordingly.

In contrast, both operator-based models remain essentially flat across the full resolution range. The FNO-SIREN backbone alone is already substantially less sensitive to $N$ than any of the step-wise baselines. This confirms that the discretization-invariance property is valid for both the purely elastoplastic case and the more complex case in which damage and plasticity play a role. Adding causal self-attention on top of the FNO-SIREN backbone yields the lowest error at every resolution and the smallest standard deviation across test trajectories, providing the most robust surrogate among the six. This confirms that the architectural advantages, i.e.\ the spectral parameterization in physical wavenumbers, the SIREN representation of sharp transitions, and the adaptive causal attention, transfer cleanly to the coupled plastic-damage setting and are not due to the simplicity of the purely elastoplastic model. 

\subsection{2D plane strain elastoplastic material model }
This section extends the evaluation to a two-dimensional plane-strain setting, where the surrogate must simultaneously predict three stress components from three strain components throughout the loading history. The added difficulty is that the trained model must learn the coupling between normal and shear components. Consequently, instead of a one-dimensional strain input, the input now consists of three strain components, forming an input tensor with three channels, as opposed to a single channel previously.

\subsubsection{Datasets}

The training set is constructed to expose the surrogate to a broad span of loading modes. Three families of Gaussian-Process strain paths are combined:

\begin{itemize}
    \item \textbf{Multiaxial strain condition GP (10000 paths):} every channel/dimension is an independently drawn Gaussian-Process path with range from $[-1,1]$, anchored to $\boldsymbol{\varepsilon}(0) = \mathbf{0}$. These paths exercise all three components simultaneously.
    \item \textbf{Uniaxial strain condition GP (2000 paths):} only a single randomly chosen channel/dimension is active; the other two channels/dimensions are identically zero. These paths exercise pure normal-only or pure shear-only loading.
    \item \textbf{Biaxial strain condition GP (2000 paths):} two randomly chosen channels/dimensions are active, the third is zero. These paths exercise two-component combinations.
\end{itemize}

The combined training pool of 14000 paths is split 90/10 into training and validation. All paths use $N = 50$ sampling points and start at the undeformed state. 

\subsubsection{Representative predictions}
\label{sec:representative predictions 2D}
\begin{figure}[H]
  \centering
  
  \begin{subfigure}[b]{0.99\linewidth}
    \centering
    \includegraphics{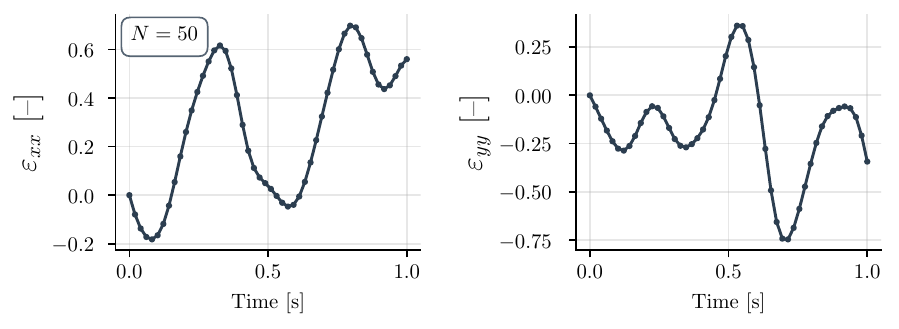}
    \subcaption{Strain loading path at training resolution}
    \label{fig:test_in_temp50_2d}
  \end{subfigure}
\end{figure}

To illustrate the qualitative behavior of the RNN and proposed material operator for the 2D elastoplastic material model before 
presenting aggregate statistics, we examine representative stress-strain 
hysteresis loops under two evaluation conditions. 
In each case we compare the RNN-GRU surrogate, and the proposed 
FNO-SIREN with causal attention operator. The two conditions are 
(i)~a Gaussian-Process biaxial loading path evaluated at the training 
resolution $N=50$ as shown in Figure \ref{fig:test_in_temp50_2d}, which serves as a sanity check; 
(ii) a complex, variable-amplitude cyclic biaxial loading path (evaluated at $N=1000$) as shown in Figure \ref{fig:test_out_temp50_2d}, designed to simultaneously test path-family generalization and resolution invariance. All models are trained exclusively 
at $N=50$ on GP paths and receive no further training between conditions.

\begin{figure}[H]
  \centering
  \continuedfloat

  \begin{subfigure}[b]{0.99\linewidth}
    \centering
    \includegraphics{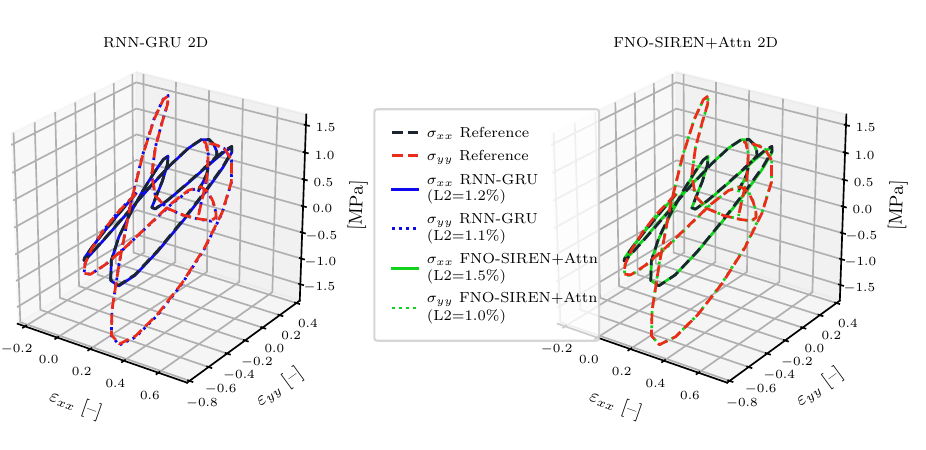}
  \end{subfigure}

  \caption{In-distribution(Gaussian-Process) loading path at temporal resolution N = 50} 
  \label{fig:in_distri_temp50}
\end{figure}

\begin{figure}[H]
  \centering
  
  \begin{subfigure}[b]{0.99\linewidth}
    \centering
    \includegraphics{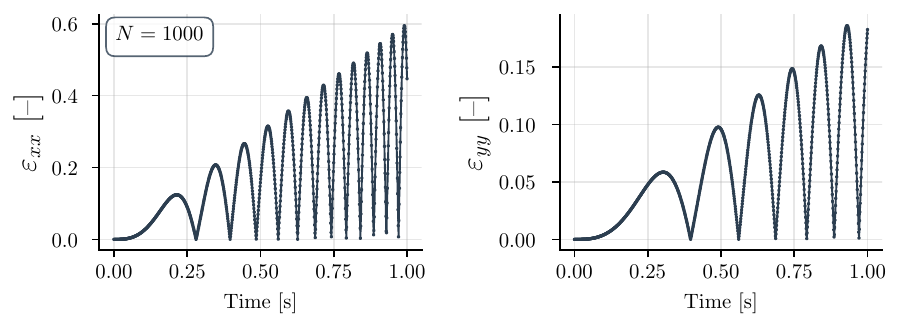}
    \subcaption{Strain loading path at temporal resolution N = 1000}
    \label{fig:test_out_temp50_2d}
  \end{subfigure}

  \begin{subfigure}[b]{0.99\linewidth}
    \centering
    \includegraphics{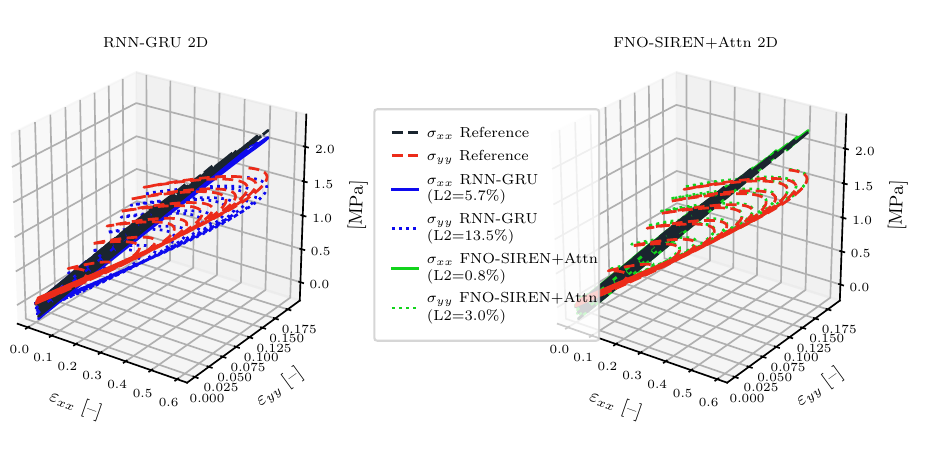}
  \end{subfigure}

 \caption{Complex loading path at temporal resolution N = 1000 }
  \label{fig:out_distri_temp1000}
\end{figure}

Figure \ref{fig:in_distri_temp50} shows that, at the training resolution, the RNN-GRU model and the proposed operator model predict quite accurately on paths similar to the Gaussian training paths. Reference solutions are obtained from the numerically calculated material model. Model accuracy is evaluated using the relative L2 error. All models had relative error with respect to the reference solution in the range of 0-2\%.

However, when evaluated at a higher resolution with more complex loading paths, as shown in Figure \ref{fig:out_distri_temp1000}, the RNN-GRU model shows an increase in error of up to 13.5\% for some components, while the proposed operator architecture, FNO-SIREN+Attention, had substantially lower error in the range of 0.8\%-3.0\%. To further investigate this behavior, we conducted a statistical study to assess discretization invariance. 

\subsubsection{Discretization study}

We repeat the discretization-invariance test, now on the 100 canonical zig-zag  loading paths consisting of uniaxial, biaxial, and multiaxial loading. The model is trained only at $N=50$ and evaluated at finer resolutions. Figure \ref{fig:discretization_study_2d} reports the mean relative $L_2$ error and its standard deviation band for RNN-GRU, FNO-SIREN and FNO-SIREN+Attention models .

\begin{figure}[H]
  \centering
  \begin{subfigure}{\textwidth}
    \centering
    \includegraphics{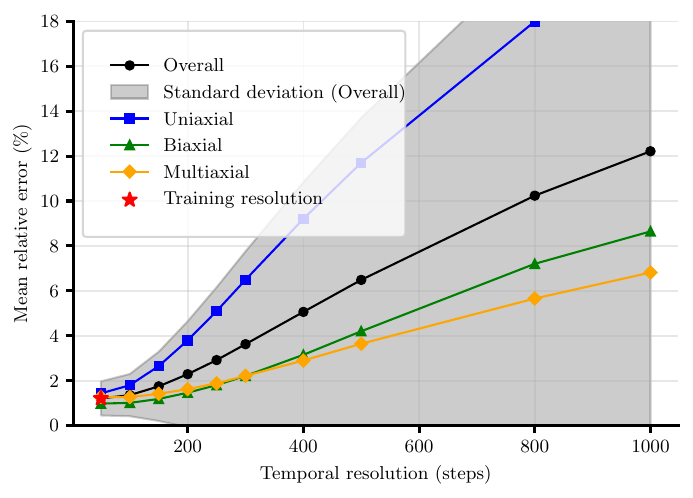}
    \caption{RNN-GRU}
    \label{fig:t1}
  \end{subfigure}
\end{figure}

\begin{figure}[H]
  \ContinuedFloat
  \centering
  \begin{subfigure}{\textwidth}
    \centering
    \includegraphics{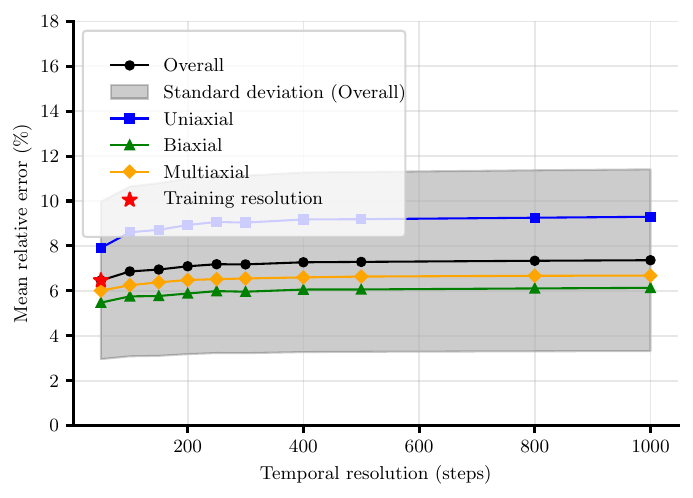}
    \caption{FNO-SIREN}
    \label{fig:t2}
  \end{subfigure}
\end{figure}

\begin{figure}[H]
  \ContinuedFloat
  \centering
  \begin{subfigure}{\textwidth}
    \centering
    \includegraphics{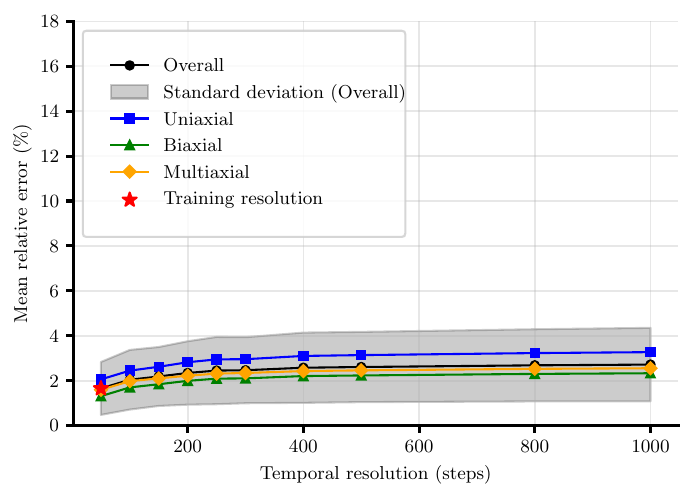}
    \caption{FNO-SIREN+Attention}
    \label{fig:t3}
  \end{subfigure}
\end{figure}

\begin{figure}[H]
  \ContinuedFloat
  \centering
  \begin{subfigure}{\textwidth}
    \centering
    \includegraphics{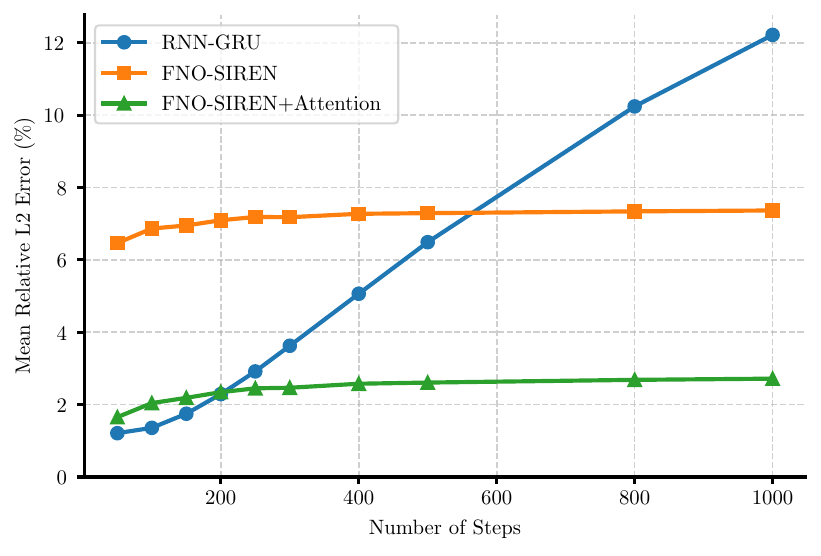}
    \caption{Overall mean error for all architectures}
    \label{fig:t4}
  \end{subfigure}
  
  \vspace{1em} 
  
  \caption{Discretization study comparing RNN-GRU, FNO-SIREN, and FNO-SIREN with Attention.  }
  \label{fig:discretization_study_2d}
\end{figure}

For FNO-SIREN and FNO-SIREN+Attention, the overall mean error remains almost constant (flat) across the entire range of sampling resolutions. Each type of loading (uniaxial, biaxial or multiaxial) curve exhibits the same (constant error) flat behavior, all predicted with comparable or better accuracy regardless of the temporal sampling density. The standard-deviation band remains narrow throughout, indicating that the discretization invariance is not the result of averaging over a heterogeneous distribution of per-trajectory errors but a structural property that holds path-by-path. These results confirm the successful generalization of the architectural ingredients identified in previous sections. Specifically, spectral parameterization indexed by physical wavenumbers, SIREN representation of sharp transitions, and adaptive causal self-attention transition seamlessly from a scalar one-dimensional to the two-dimensional case. The operator learned at $N=50$ delivers predictions of comparable quality at $N=1000$ across loadings (uniaxial, biaxial, or multiaxial).

\section{Accuracy and computational efficiency of surrogate models}

We compare all six surrogate models in terms of predictive accuracy 
and computational cost. Accuracy is reported at both the training 
resolution $N = 50$ and the highest evaluated resolution $N = 1000$, 
so that the table captures both the in-distribution performance and 
the resolution-generalization behavior simultaneously. 
Sections~\ref{sec:representative predictions 1D elasto},~\ref{sec:representative predictions 1D ductile-damage}, and~\ref{sec:representative predictions 2D} present the predicted 
stress--strain responses for representative loading paths under the 
three evaluation conditions. Table~\ref{tab:comparison table} 
summarizes the mean relative $L_2$ error over 100 zig-zag test paths 
for the 1D elastoplasticity benchmark, together with training and 
per-trajectory inference times. Data generation is shared across 
all models; this cost is excluded from the 
per-model training times reported below.

\begin{table}[H]
    \centering
    \begin{tabularx}{\textwidth}{|>{\centering\arraybackslash}X|>{\centering\arraybackslash}X|>{\centering\arraybackslash}X|>{\centering\arraybackslash}X|>{\centering\arraybackslash}X|} 
        \hline
        \textbf{Model Architectures} & \textbf{Mean Error at training resolution (\%)} & \textbf{Mean Error at $N = 1000$ resolution (\%)} & \textbf{Training Time  (min)} & \textbf{Inference time per signal (msec)}\\ \hline
        FFNN (W=1) & 1.51 & 13.27 & \underline{14.6} & 12.882 \\ \hline
        FFNN (W=5) & 1.29 & 13.83 & \textbf{10.8} & 14.935 \\ \hline
        FFNN (W=10) & 1.2 & 9.87 & 24.1 & 14.891 \\ \hline
        RNN-GRU & \textbf{0.32} & 12.41 & 17.28 & 18.629 \\ \hline
        FNO-SIREN & 4.28 & \underline{7.49} & 22.36 & \textbf{1.487} \\ \hline
        FNO-SIREN+\allowbreak Attention & \underline{0.75} & \textbf{2.01} &103.45 & \underline{2.617} \\ \hline
    \end{tabularx}
    \caption{Performance comparison between various model architectures and the traditional return mapping algorithm (baseline inference time: 51.156 ms) for the 1D elastoplasticity material model. All metrics were benchmarked on an NVIDIA GeForce RTX 4090 GPU.}
    \label{tab:comparison table}
\end{table}

The results summarized in Table \ref{tab:comparison table} reveal several important trends regarding the trade-offs between accuracy, generalization, and computational cost across the six surrogate architectures evaluated. For each metric, the best and second-best results are highlighted in boldface and underlined, respectively.

Across the six surrogate architectures, RNN-GRU achieves the best accuracy at the training resolution ($0.32\%$), reflecting its natural suitability for capturing the path-dependent, history-sensitive nature of elastoplastic loading. However, this advantage does not carry over to the higher evaluation resolution ($N = 1000$), where RNN-GRU's error rises sharply to $12.41\%$, similar to the degradation seen in the FFNN variants ($9.87$--$13.83\%$). This indicates that purely feed-forward and recurrent architectures, while effective within their training discretization, lack the inductive bias needed to generalize across resolutions. In contrast, the FNO-based models generalize considerably better as FNO-SIREN leads to $7.49\%$ error at $N = 1000$. The FNO-SIREN+Attention model achieves the best overall balance, with $0.75\%$ error at training resolution and only $2.01\%$ at $N = 1000$ which shows an order of magnitude better generalization than the FFNN and RNN-GRU models, consistent with the resolution-invariance properties of Fourier-based operator learning enhanced by attention.

These accuracy gains come with trade-offs in computational cost. The FNO-SIREN+Attention surrogate requires by far the longest training time ($103.45$~min, roughly $4$--$9\times$ the other models), reflecting the added complexity of combining spectral convolutions with attention. Despite this, its inference time ($2.617$~msec/signal) is much faster than the FFNN and RNN-GRU models ($12.8$--$18.6$~msec/signal), whose sequential or dense computations are less efficient at inference. Overall, FNO-SIREN+Attention is the preferred choice when resolution-robust predictions and fast inference are required, justifying its higher one-time training cost for deployment scenarios involving varying discretizations. While not applied to the 1D coupled plastic-damage model, FNO-SIREN+Attention model should yield equal or greater speed-ups than the elastoplastic case due to its reduced layer and Fourier mode requirements (Table \ref{tab:hyperparameters}). Instead, we extended the study to the multidimensional case. Results are presented in Table \ref{tab:comparison table 2d}. The observed trends remain consistent with those reported above.

\begin{table}[H]
    \centering
    \begin{tabularx}{\textwidth}{|>{\centering\arraybackslash}X|>{\centering\arraybackslash}X|>{\centering\arraybackslash}X|>{\centering\arraybackslash}X|>{\centering\arraybackslash}X|} 
        \hline
        \textbf{Model Architectures} & \textbf{Mean Error at training resolution (\%)} & \textbf{Mean Error at $N = 1000$ resolution (\%)} & \textbf{Training Time  (min)} & \textbf{Inference time per signal (msec)}\\ \hline
        FFNN (W=1) & \textbf{1.201} & 12.502 & \underline{29.82} & 12.267 \\ \hline
        FFNN (W=5) & 1.453 & 15.671 & 49.15 & 16.897 \\ \hline
        FFNN (W=10) & 1.425 & 18.273 & 38.82 & 16.988 \\ \hline
        RNN-GRU & \underline{1.211} & 12.215 & \textbf{14.38} & 18.503 \\ \hline
        FNO-SIREN & 6.463 & \underline{7.364} & 197.42 & \textbf{1.537} \\ \hline
        FNO-SIREN+\allowbreak Attention & 1.654 & \textbf{2.717} & 273.40 & \underline{2.827} \\ \hline
    \end{tabularx}
    \caption{Performance comparison between various model architectures and the traditional return mapping algorithm (baseline inference time: 30.470 ms) for the 2D elastoplasticity material model. All metrics were benchmarked on an NVIDIA GeForce RTX 4090 GPU.}
    \label{tab:comparison table 2d}
\end{table}

\begin{figure}[H]
  \centering
  
  \begin{subfigure}[b]{0.49\linewidth}
    \centering
    \includegraphics{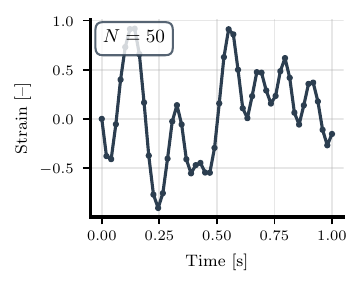} 
    \subcaption{In-distribution loading path at higher resolution}
  \end{subfigure}
  \hfill
  \begin{subfigure}[b]{0.49\linewidth}
    \centering
    \includegraphics{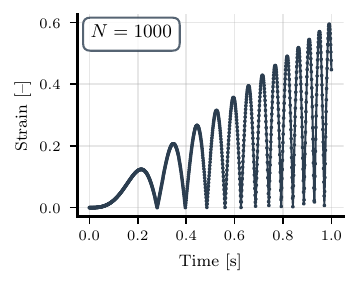} 
    \subcaption{Complex loading path at higher resolution}
  \end{subfigure}
\vfill
  \begin{subfigure}[b]{0.49\linewidth}
      \centering
      \includegraphics{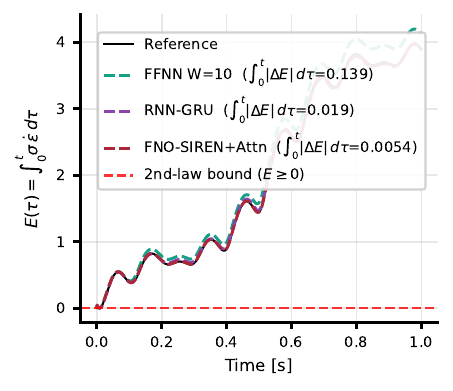}
      \subcaption{Mechanical work for 1D elastoplasticity material model}

  \end{subfigure}
  \hfill
  \begin{subfigure}[b]{0.49\linewidth}
      \centering
      \includegraphics{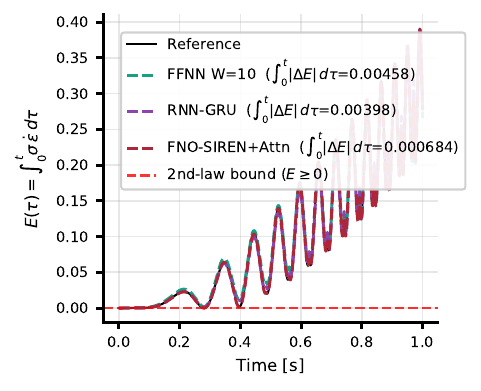}
      \subcaption{Mechanical work for 1D elastoplasticity material model }
  \end{subfigure}

  \begin{subfigure}[b]{0.49\linewidth}
      \centering
      \includegraphics{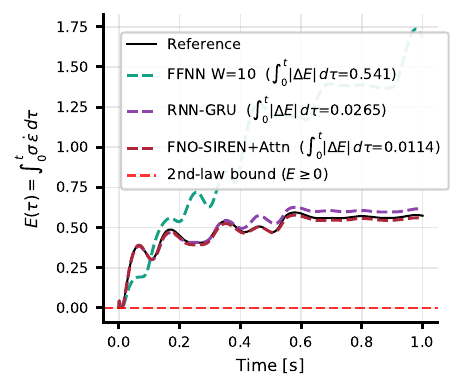}
      \subcaption{Mechanical work for 1D coupled damage-plasticity model}

  \end{subfigure}
  \hfill
  \begin{subfigure}[b]{0.49\linewidth}
      \centering
      \includegraphics{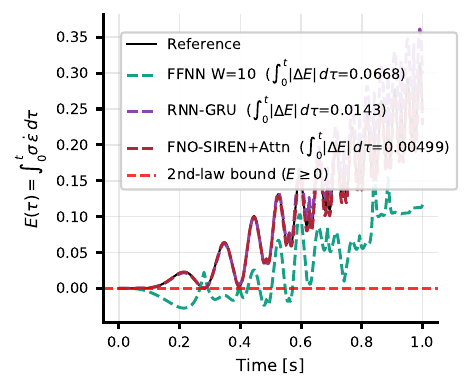}
      \subcaption{Mechanical work for 1D coupled damage-plasticity model}

  \end{subfigure}


  \caption{Mechanical work difference between surrogate models and reference models  } 
  \label{fig:diss_err}
\end{figure}

Figure~\ref{fig:diss_err} evaluates the extent to which the trained surrogate models are thermodynamically consistent with the Newton-Raphson reference, by measuring the discrepancy between their implied cumulative mechanical work over time. For a stress history $\boldsymbol{\sigma}(t)$ produced along an applied strain path $\boldsymbol{\varepsilon}(t)$, the cumulative mechanical work done by the stress up to time $t$ is
\begin{equation}
    E(t) = \int_0^t \boldsymbol{\sigma} : \dot{\boldsymbol{\varepsilon}} \, d\tau.
\end{equation}

The local Clausius--Duhem inequality states that the mechanical dissipation ${D} = \boldsymbol{\sigma} : \dot{\boldsymbol{\varepsilon}} - \dot{\psi}$ must be non-negative at every instant, where $\psi$ is the Helmholtz free energy. Integrating this pointwise statement from $0$ to $t$ gives the time-integrated form:

\begin{equation}
    \int_{0}^{t} {D} \, \mathrm{d}\tau = \int_{0}^{t} \left( \boldsymbol{\sigma} : \dot{\boldsymbol{\varepsilon}} - \dot{\psi} \right) \mathrm{d}\tau \geq 0 \quad \rightarrow \quad D(t) = E(t) - \psi(t) \geq 0
\end{equation}

Evaluating $\psi(t)$ in general requires knowledge of the plastic hardening and 
damage contributions to the free energy, in addition to the elastic part. 
However, the surrogate models are trained purely on strain-to-stress data, 
without access to any constitutive parameters or internal variables, so $\psi(t)$ 
itself is not directly accessible. We therefore make the reasonable assumptions 
that the initial state of the material coincides with its natural, virgin state, 
i.e., $\psi(0) = 0$, $D(t=0)=0$ and that the Helmholtz free energy is non-negative for all 
time, $\psi(t) \geq 0$.

Under these assumptions, $E(t)$ can serve as a 
model-agnostic, thermodynamically admissible quantity that is checkable directly 
from the predicted stress--strain response. Since $D(t) \geq 0$ and $\psi(t) \geq 0$, it follows that $E(t) = D(t) + \psi(t) \geq 0$ holds for every material for any thermodynamically admissible process. A surrogate whose predicted $E(t)$ deviates from this violates the second law of thermodynamics.

We initially attempted to exploit this same conservativeness during training, imposing $E(t) \geq 0$ directly as a penalty on the network's predictions. This did not produce a significant change in the predictive behavior of the surrogates. Because the reference solution satisfies $E(t) \geq 0$ exactly by construction as shown in Figure \ref{fig:diss_err}, we quantify each surrogate's departure from thermodynamic consistency:

\begin{equation}
    \int_0^t |\Delta E| \, d\tau = \int_0^t \left| E_{\text{pred}}(\tau) - E_{\text{ref}}(\tau) \right| \, d\tau
\end{equation}

At higher resolution, all three evaluated models show a comparatively  small accumulated error, with the FNO-SIREN+Attention model showing the smallest accumulated error, followed by RNN-GRU and then FFNN. This indicates that FNO-SIREN+Attention's stress predictions track the reference's energy accurately under  high resolution and complex paths.

\section{Potential extension to FE solver}

A question for any data-driven constitutive surrogate is whether it can be embedded in a finite element solver, where the constitutive law is queried incrementally. The internal variables that a classical model would carry between steps are, in our case, never predicted or calculated at all. Since the proposed operator has no explicit internal state, it is not obvious a priori that it can be used in this incremental setting without retraining or redesign.

\begin{figure}[H]
  \centering
  
  \begin{subfigure}[b]{0.49\linewidth}
    \centering
    \includegraphics{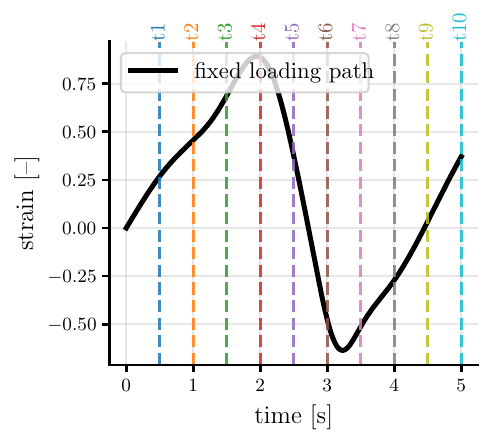}
    \subcaption{Input strain}
    \label{fig:input_strain}
  \end{subfigure}
  \hfill
  \begin{subfigure}[b]{0.49\linewidth}
    \centering
    \includegraphics{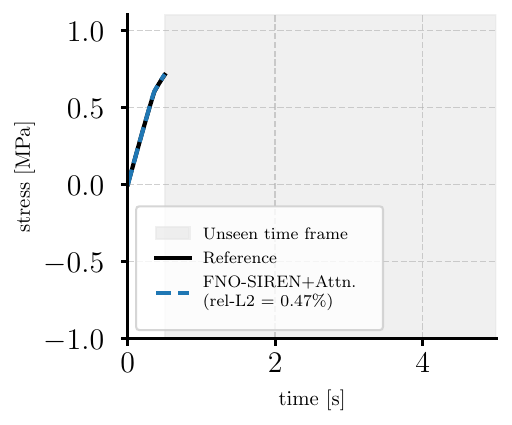}
    \subcaption{Stress field till t1 (0.5 s)}
    \label{fig:0.5}
  \end{subfigure}
 \begin{subfigure}[b]{0.49\linewidth}
    \centering
    \includegraphics{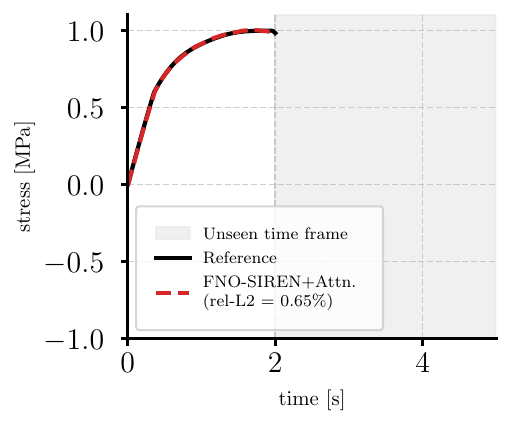} 
    \subcaption{Stress field till t4 (2 s)}
    \label{fig:2.0}
  \end{subfigure}
  \hfill
  \begin{subfigure}[b]{0.49\linewidth}
    \centering
    \includegraphics{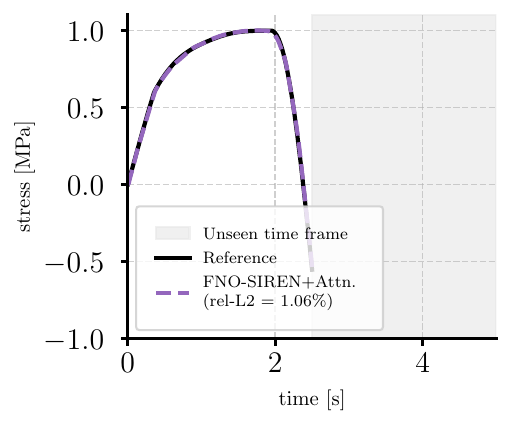} 
    \subcaption{Stress field till t5 (2.5 s)}
    \label{fig:2.5}
  \end{subfigure}

  \begin{subfigure}[b]{0.49\linewidth}
    \centering
    \includegraphics{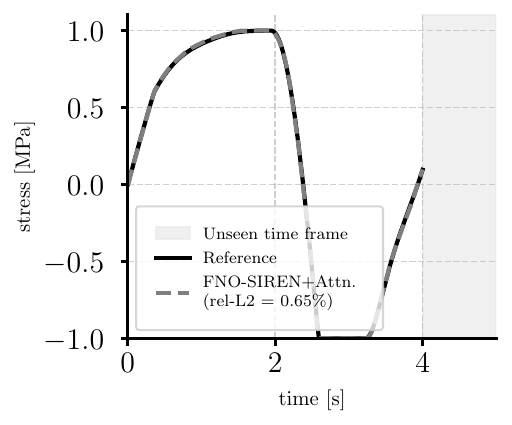} 
    \subcaption{Stress field till t8 (4 s)}
    \label{fig:4.0}
  \end{subfigure}
  \hfill
  \begin{subfigure}[b]{0.49\linewidth}
    \centering
    \includegraphics{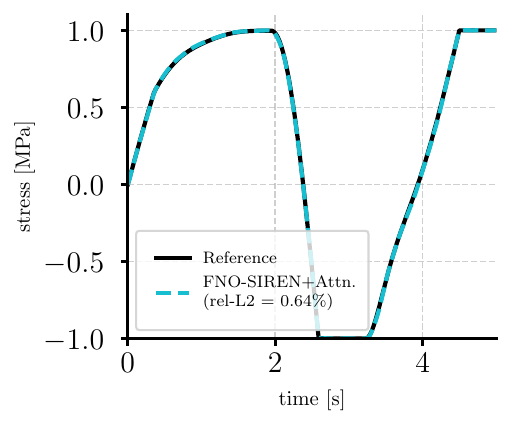} 
    \subcaption{Stress field till t10 (5 s)}
    \label{fig:5.0}
  \end{subfigure}

  \caption{No error accumulation across independently evaluated horizons }
  \label{fig:fe_windows}
\end{figure}

To test this, we take a single loading path and query the trained operator at increasing horizons, corresponding to the path truncated at $t = 0.5\text{s}, 1.0\text{s}, \dots, 5.0\text{s}$. At each horizon, the model is given only the strain history up to that point and predicts the stress over that same interval, as an FE solver would present a growing loading history to the constitutive law as the simulation advances. Figure~\ref{fig:fe_windows} shows the input strain path together with five representative horizons against the reference return-mapping solution for the 1D elastoplastic material model (Section~\ref{subsec: 1-d elasto}); the remaining horizons follow the same trend and are omitted for space.

Two observations follow directly from these results. First, the model produces an accurate stress response at every horizon, with the relative error remaining between 0.33\% and 1.06\% across all ten windows and showing no upward trend as the horizon lengthens. Second, and more significant for FE deployment, each prediction is an independent forward pass rather than an extension of the previous one. It is recomputed from scratch from the full strain history up to that point. If predictions at later horizons were built on top of earlier ones, any small inaccuracy at t = 0.5s would compound as the horizon grows, and this doesn't happen in our case as shown in Figure \ref{fig:fe_windows}. This points to a practical strategy for coupling the operator to an FE solver. At each global time step, the solver would supply the strain history recorded so far at a given integration point, the operator would return the full stress history over that interval, and only the stress at the current time step would be retained for the equilibrium iteration, with the rest of the returned trajectory discarded. Because each call is independent, this requires no internal state to be tracked by the solver between steps beyond the strain history itself. The results in Figure~\ref{fig:fe_windows} indicate that this repeated-evaluation strategy would remain accurate throughout the simulation, since the operator's accuracy at a given horizon does not depend on how many times it has been queried at earlier horizons.

\section{Conclusion and Outlook}

This work addressed a structural limitation that has long constrained data-driven constitutive modeling of rate-independent path-dependent materials, which is the tying of the learned representation to the temporal discretization of the training data. Step-wise surrogates, such as feedforward networks with sliding history windows and recurrent networks with hidden-state memory, process the loading path one increment at a time, and their predictions degrade when the test loading path is sampled at a resolution different from the training resolution. For a rate-independent constitutive law, whose response should depend only on the loading path and not on how it is sampled, this is a mechanical inconsistency.

We proposed a sequence-to-sequence operator that departs from both autoregressive and sequential-recurrent paradigms. Instead of consuming the loading path incrementally, the proposed model presents the entire discretely sampled strain trajectory as a single function input and produces the entire stress trajectory in one parallel forward pass. Path dependence is architecturally encouraged by a strict causal mask inside every self-attention layer. The architecture combines three complementary ingredients: an FNO backbone for resolution-invariant spectral convolution, sinusoidal representation layers (SIREN) for the high-fidelity recovery of sharp transitions characteristic of yield onset, unloading, and damage initiation, and causal multi-head self-attention for adaptive weighting of the strain history.

We evaluated the framework using three constitutive models of increasing complexity. We began with one-dimensional elastoplasticity featuring nonlinear isotropic hardening, moved on to one-dimensional coupled damage-plasticity, and concluded with two-dimensional plane-strain $J_2$ elastoplasticity. Across all three benchmarks the proposed model achieved quite low prediction error and the smallest sensitivity to the temporal discretization, retaining essentially flat error curves across a twenty-fold change in sampling resolution from $N=50$ (the training resolution) up to $N=1000$. The operator achieved its best performance when the internal variables were highly coupled, as demonstrated in the coupled damage-plasticity model. In this scenario, the proposed framework showed the largest advantage over the baseline data-driven approaches. Furthermore, the core architectural features enabling this robustness transferred without modification to both the coupled damage-plasticity and multi-component plane-strain settings. In the latter case, the operator successfully learned the coupling between normal and shear components alongside the differential evolution of the plastic variables. In terms of computational efficiency, the proposed operator also delivered substantially faster inference than competing data-driven approaches, despite its added architectural complexity. This speed advantage, given a one-time training cost per material model, makes the framework particularly well suited for deployment scenarios requiring repeated or large-scale evaluation.

The observed discretization invariance has a simple interpretation. The model has learned the continuous strain-to-stress mapping itself, not an approximation tied to one step size. This is exactly the behavior a rate-independent constitutive law requires, and in the proposed framework it follows directly from the architecture, without any additional constraint or penalty during training. 

The present study is limited to rate-independent materials, for which the response depends on the loading path but not on the loading rate. Extending the framework to viscoelastic and viscoplastic materials requires the temporal axis to carry physical-time information rather than to act purely as a sample index, which in turn requires the spectral filters to be parameterized in terms of true frequency rather than dimensionless wavenumber. This extension is conceptually compatible with the proposed architecture and would broaden its applicability to a much wider class of inelastic materials. The current architecture already encourages causality in the material operator through the causal attention mask. Building on this, a natural next step is to additionally enforce thermodynamic admissibility. An additional extension can be to condition the operator on a microstructural descriptor. The resulting joint space-time operator would predict the full stress response under arbitrary loading from a heterogeneous microstructure, providing a unified surrogate for both the constitutive behavior and the microstructural variability.

Together, these future directions aim to use the current discretization-invariant surrogate to produce a physics-consistent and microstructure-aware operator, fully integrated within finite element frameworks for complex inelastic materials.

\section*{Data Availability}
All codes used for this study are openly available in folax at \href{https://github.com/Neural-Mechanics-Lab/folax/tree/Material_operator/examples/Material_operator}{Material-Operator}.

\section*{Acknowledgements}
R. Arora, T. Brepols, and S. Rezaei thankfully acknowledge the funding of the Deutsche Forschungsgemeinschaft (DFG, German Research Foundation) – Project number 561202254. L. Scheunemann gratefully acknowledges the funding of project SPP 2489 (project number 562153065) by the DFG. L. Scheunemann and T. Brepols gratefully acknowledge the funding of project CRC/TRR 339 (subproject B05, project number 453596084) by the DFG. During the preparation of this work, the authors used Claude (Opus 4.8, Anthropic) to assist with drafting and editing portions of the manuscript text. After using this tool/service, the authors reviewed and edited the content as needed and take full responsibility for the content of the published article.

\section*{Funding Information}
This work was supported by the Deutsche Forschungsgemeinschaft (561202254, 562153065, 453596084).

\section*{Conflict of Interest Statement}
The authors declare no conflict of interest.

\section*{Appendix A. Discretization study on sinusoidal loading paths}
\begin{figure}[H]
    \centering
    \includegraphics{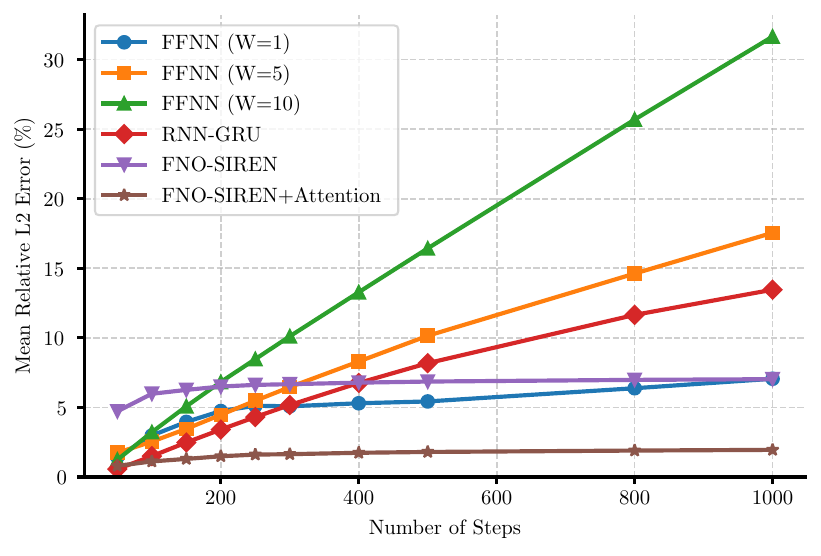}
    \caption{Discretization study on sinusoidal loading paths, 1D elastoplastic material model.}
    \label{fig:A1}
\end{figure}

\begin{figure}[H]
    \centering
    \includegraphics{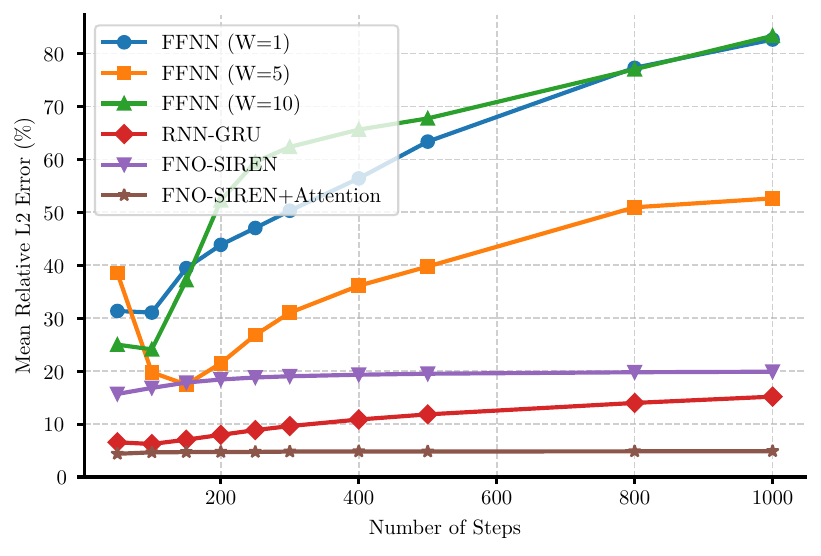}
    \caption{Discretization study on sinusoidal loading paths, 1D coupled damage-plasticity model.}
    \label{fig:A2}
\end{figure}

\begin{figure}[H]
    \centering
    \includegraphics{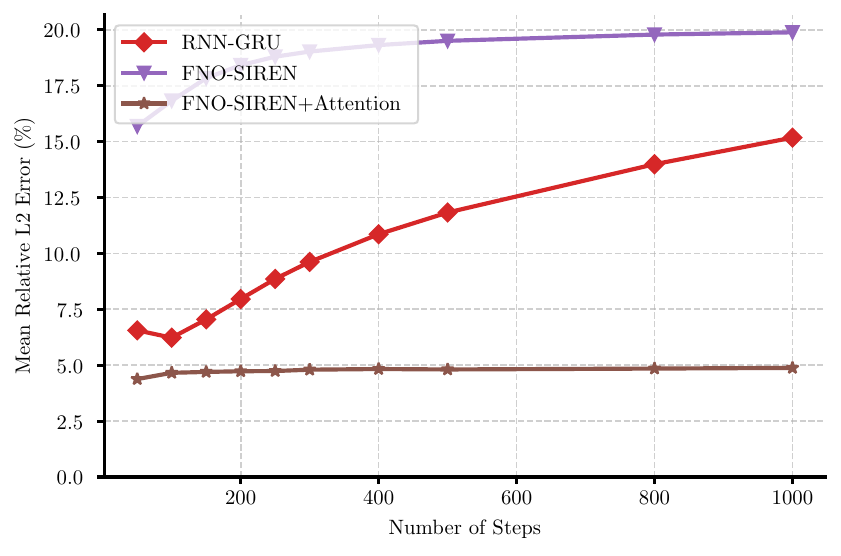}
    \caption{Discretization study on sinusoidal loading paths, 1D coupled damage-plasticity model.}
    \label{fig:A3}
\end{figure}

We repeat the procedure of Sections~\ref{subsec: Discretization_study_1d_elasto} and \ref{subsec: Discretization_study_1d_damage} on the 100 canonical sinusoidal test paths, evaluating all six models (trained at $N = 50$) at resolutions $N \in \{50$, $100$, $150$, $200$, $250$, $300$, $400$, $500$, $800$, $1000\}$. Figures~\ref{fig:A1}, \ref{fig:A2} and \ref{fig:A3} report the mean relative $L_2$ error as a function of temporal resolution. The overall picture is consistent with the zig-zag study: the FFNN($W=1$), FFNN($W=5$), FFNN($W=10$), and RNN-GRU models all degrade steadily as resolution increases. The FNO-SIREN model starts from a higher baseline error but remains comparatively flat across the full resolution range, confirming that the discretization-invariance of the spectral backbone holds independently of the attention mechanism. Adding attention improves accuracy substantially without sacrificing this discretization-invariant behavior.


\clearpage
\newpage
\bibliography{Ref}
\end{document}